\def\year{2021}\relax
\documentclass[letterpaper]{article} 
\usepackage{aaai21}  
\usepackage{times}  
\usepackage{helvet} 
\usepackage{courier}  
\usepackage[hyphens]{url}  
\usepackage{graphicx} 
\usepackage{graphicx}  
\usepackage{natbib}  
\usepackage{caption} 
\usepackage{amssymb}
\usepackage{mathrsfs}
\usepackage{amsmath}
\usepackage{mathtools}
\usepackage{resizegather}

\title{Understanding the Role of Adversarial Regularization in Supervised Learning}

\author {
	Litu Rout\footnote{Under Review} \\
}

\affiliations{Space Applications Centre \\
	Indian Space Research Organization\\
	lr@sac.isro.gov.in
}

\begin{document}
	\maketitle
	
	\begin{abstract}
		Despite numerous attempts sought to provide empirical evidence of adversarial regularization outperforming sole supervision, the theoretical understanding of such phenomena remains elusive. In this study, we aim to resolve whether adversarial regularization indeed performs better than sole supervision at a fundamental level. To bring this insight into fruition, we study vanishing gradient issue, asymptotic iteration complexity, gradient flow and provable convergence in the context of sole supervision and adversarial regularization. The key ingredient is a theoretical justification supported by empirical evidence of adversarial acceleration in gradient descent. In addition, motivated by a recently introduced unit-wise capacity based generalization bound, we analyze the generalization error in adversarial framework. Guided by our observation, we cast doubts on the ability of this measure to explain generalization. We therefore leave as open questions to explore new measures that can explain generalization behavior in adversarial learning. Furthermore, we observe an intriguing phenomenon in the neural embedded vector space while contrasting adversarial learning with sole supervision. 
	\end{abstract}
	
	\section{Introduction}
	At a fundamental level, we study the role of adversarial regularization in supervised learning through the lens of theoretical justification. We intend to resolve the mystery of why supervised learning with adversarial regularization accelerates gradient updates as compared to sole supervision. In light of deeper understanding, we explore several crucial properties pertaining to adversarial acceleration in gradient descent.
	
	Over the years several variants of gradient descent algorithms have emerged. In various tasks, adaptive methods including Adagrad~\cite{duchi2011adaptive}, Adadelta~\cite{zeiler2012adadelta}, RMSProp~\cite{tieleman2012lecture}, ADAM~\cite{kingma2014adam}, and NADAM~\cite{dozat2016incorporating} perform relatively better than classical gradient descent. Of particular interest, stochastic  version of gradient descent, namely SGD with momentum has enjoyed great success in neural network optimization. Its simplicity, superior performance~\cite{wilson2017marginal}, and theoretical guarantees~\cite{carmon2018accelerated} often provide an edge over other contemporary learning algorithms in several tasks. For this reason, we choose SGD as our primary learning algorithm to foster smooth transition from recent analyses~\cite{nagarajan2017generalization,neyshabur2019role}. We argue that despite superior performance, it suffers from vanishing gradient issue in near optimal region. In fact, this is mirrored by poor practical performance when compared with adversarial regularization as independently reported in copious literature~\cite{denton2015deep,wang2016generative,ledig2017photo,rangnekar2017aerial,wang2018esrgan,xue2018segan,xian2018texturegan}. We identify the root cause of this issue to be the primary objective function. Since these methods rely on some form of gradients estimated from the supervised objective, the issue of vanishing gradient inherently resides in near optimal region.

	In recent years, the research community has witnessed pervasive use of Generative Adversarial Networks (GANs) on a wide variety of complex tasks~\cite{isola2017image,zhu2017unpaired,park2019semantic,karras2019style}. Among many applications, some require generation of a particular sample subject to a conditional input. For this reason, there has been a surge in designing conditional adversarial networks~\cite{mirza2014conditional}. In visual object tracking via adversarial learning, Euclidean norm is used to regulate the generation process so that the generated mask falls within a small neighborhood of actual mask~\cite{song2018vital}. In photo-realistic image super resolution, Euclidean or supremum norm is used to minimize the distance between reconstructed and original image~\cite{ledig2017photo,wang2018esrgan}. In medical image segmentation, multi-scale $L_1$-loss with adversarial regularization is shown to outperform sole supervision~\cite{xue2018segan}. In medical image analysis, a 3d conditional GAN along with $L_1$-distance is used to super resolve CT scan imagery~\cite{kudo2019virtual}.
	
	Furthermore, \citeauthor{isola2017image} use $L_1$-loss as a supervision signal and adversarial regularization as a continuously evolving loss function. Because GANs learn a loss that adapts to data, they fairly solve multitude of tasks that would otherwise require hand-engineered loss. \citeauthor{xian2018texturegan} use adversarial loss on top of pixel, style, and feature loss to restrict the generated images on a manifold of real data. Prior works on this operate under the synonym conditional GAN where a convex composition of pixel and adversarial loss is primarily optimized~\cite{mirza2014conditional,denton2015deep,wang2016generative}. \citeauthor{karacan2016learning} use this technique to efficiently generate images of outdoor scenes. \citeauthor{rout2020s2a} combine spatial and Laplacian spectral channel attention in regularized adversarial learning to synthesize high resolution images. \citeauthor{emami2019spa} coalesce spatial attention with adversarial regularization and feature map loss to perform image-to-image translation. 
	
	As per these prior and concurrent works~\cite{rangnekar2017aerial,xue2018segan,rout2020alert,dong2015image,henaff2019model,sarmad2019rl}, it is understandable that supervised learning with adversarial regularization boosts empirical performance. More importantly, this behavior is consistent across a wide variety of problems and network configurations. As much beneficial as this has been so far, to our knowledge, the theoretical understanding still remains relatively less explored. Aiming to bridge this gap, we provide theoretical and empirical evidence of better performance due to adversarial regularization when compared with sole supervision.
	
	\section{Related Works}
	\subsubsection{Adversarial Regularization}
	The spectral and spatial super resolution based on adversarial regularization~\cite{rangnekar2017aerial,rout2020alert} is proven to achieve \textit{faster convergence} and \textit{better empirical risk} compared to purely supervised learning~\cite{lanaras2018super}. Further, \citeauthor{ledig2017photo} showed improvement in perceptual quality of high resolution images in adversarial setting. Despite superior empirical performance, the theoretical understanding of such phenomena remains elusive. To this end, the theoretical analysis suggests that there is a constant that bounds the total empirical risk above~\cite{xue2018segan}. As a result, this inhibits erroneous gradient estimation by the discriminator that apparently improves perceptual quality. However, these benign properties of loss surface do not fully explain this phenomenon at a fundamental level. The present account in this paper is intended to provide further insights to this problem.
	
	Apart from supervised and adversarial learning, the notion of adversarial regularization has also been studied in Reinforcement Learning (RL). \citeauthor{henaff2019model} use adversarial learning with expert regularization to learn a predictive policy that allows to drive in simulated dense traffic. \citeauthor{sarmad2019rl} use RL agent controlled GAN and $L_2$-distance between global feature vectors to convert noisy, partial point cloud into high-fidelity data. 
	
	\subsubsection{Accelerated Gradients}
	The idea of accelerated training has long been studied. An elegant line of research focuses on variance reduction that aims to address stochastic and finite sum problems by averaging the stochastic noise~\cite{schmidt2017minimizing,zhou2018stochastic}. Among momentum based acceleration, much theoretical progress has been made to accelerate any smooth convex optimization~\cite{nesterov2012efficiency,carmon2018accelerated}. Further, many efforts have been made towards changing the step size across iterations based on estimated gradient norm~\cite{duchi2011adaptive,staib2019escaping,zhou2018convergence}. Adversarial regularization is similar to these methods in a sense that it offers acceleration in the near optimal region.
	
	\subsubsection{Minimax Optimization}
	The seminal work of \citeauthor{neumann1928theorie} in solving the problem of minimax optimization has been a central part of game theory. Recently, a rapid increase in interest is seen to study the intrinsic properties of minimax problems. The increasing popularity owes in part to the discovery of generative adversarial networks~\cite{goodfellow2014generative}. In this paper, to focus more on the empirical success of adversarial regularization, we study a simple minimax optimization problem. However, we wish to allude some interesting line of work by~\citeauthor{lin2019gradient,lin2020near,jin2019local,mertikopoulos2018cycles} in this direction that may encourage further investigation from algorithmic point of view. It will certainly be useful to borrow some ideas from the vast literature of minimax optimization under less restrictive setting. Though it is beyond the scope of this discussion, the definition of local optimality by~\citeauthor{jin2019local} is likely to pave the way for better understanding of minimax optimization, and consequently adversarial regularization.

	\section{Preliminaries}
	\subsubsection{Notations } Let $X \subset \mathbb{R}^{d_{x}}$ and $Y \subset \mathbb{R}^{d_{y}}$ where $d_x$ and $d_y$ denote input and output dimensions, respectively. The empirical distributions of $X$ and $Y$ are denoted by $\mathcal{P}_X$ and $\mathcal{P}_Y$. Given an input $x \in X$, $f(\theta;x): \mathbb{R}^{d_{x}} \rightarrow \mathbb{R}^{d_{y}}$ is a neural network with rectified linear unit (ReLU) activation, common for both supervised and adversarial learning. Here, $\theta$ denotes the trainable parameters of the generator, $f(\theta;.)$. On the other hand, the discriminator, $g(\psi;.)$ has trainable parameters collected by $\psi$. The optimal values of these parameters are represented by $\theta^*$ and $\psi^*$. For $g: \mathbb{R}^{d_{y}}\rightarrow \mathbb{R}$, $\nabla g$ denotes its gradient and $\nabla^2 g$ denotes its Hessian. Given a vector $x$, $\left \| x \right \|$ represents its Euclidean norm. Given a matrix $M$, $\left \| M \right \|$ and $\left \| M \right \|_F$ denote its spectral and Frobenius norm, respectively.
	
	\textbf{Definition 1.} ($L$-Lipschitz) \textit{A function $f$ is $L$-Lipschitz if  $\forall \theta$, $\left \|\nabla f(\theta) \right \| \leq L$.}
	
	\textbf{Definition 2.} ($\beta$-Smoothness) \textit{A function $f$ is $\beta$-smooth if $\forall \theta$, $\left \|\nabla^2 f(\theta) \right \| \leq \beta$
	}

	\subsubsection{Problem Setup} In Wasserstein GAN (WGAN) + Gradient Penalty (GP), the generator cost function is given by
	\begin{equation}
	\arg \min_{\theta} -\mathbb{E}_{x\sim \mathcal{P}_X}\left [ g\left (\psi; f\left (\theta; x \right ) \right ) \right ]
	\end{equation}
	and the discriminator cost function,
	\begin{equation}
	\label{dis}
	\begin{split}
	\arg \min_{\psi} ~&\mathbb{E}_{x\sim \mathcal{P}_X}\left [  g\left (\psi; f\left (\theta; x \right ) \right ) \right ] - \mathbb{E}_{y\sim \mathcal{P}_Y}\left [ g\left (\psi; y \right ) \right ] \\ &+ \lambda_{GP}~\mathbb{E}_{z \sim \mathcal{P}_Z}\left [ \left ( \left \| \nabla_z g\left (\psi; z \right ) \right \| -1 \right )^2 \right ].
	\end{split}
	\end{equation}
	Here, $\mathcal{P}_{Z}$ represents the distribution over samples along the line joining samples from real and generator distribution. Unlike sole supervision, the mapping function $f_\theta(.)$ in augmented objective has access to a feedback signal from the discriminator. Thus, the optimization in supervised learning with adversarial regularization is carried out by
	\begin{equation}
	\arg \min_{\theta} \mathbb{E}_{(x,y)\sim \mathcal{P}}\left [ l\left ( f(\theta;x);y \right ) -  g\left (\psi; f\left (\theta; x \right ) \right ) \right ].
	\end{equation}
	Here, $\mathcal{P}$ denotes the joint empirical distribution over $X$ and $Y$. The discriminator cost function remains identical to Wasserstein discriminator as given by equation~(\ref{dis}).

	\section{Theoretical Analysis}
	\label{method}
	This section states the assumptions and their justifications in the context of adversarial regularization. The theoretical findings are intended to provide convincing reasons to multitude of tasks that owe the benefits to adversarial training. The technical overview begins with vanishing gradient issue in the near optimal region. It then presents the main results of this study. The bounds may appear weak to some readers, but note that the goal of this study is not to provide a tighter bound individually for sole supervision and adversarial regularization. Rather, the goal is to understand the role of adversarial regularization in supervised learning --- whether adversarial regularization helps tighten the existing bounds in supervised learning literature. Thus, the emphasis is on providing a theoretical justification to the practial success of supervised learning with adversarial regularization.
	
	\subsection{Warm-Up: Mitigating Vanishing Gradient in Near Optimal Region}
	The primary assumptions are stated as following.
	
	\textbf{Assumption 1.} \textit{The mapping function $f(\theta;x)$ is $L$-Lipschitz in $\theta$.}
	
	\textbf{Assumption 2.} \textit{The loss function $l(p;y)$, where $p=f(\theta;x)$, is $\beta$-smooth in $p$.}
	
	\textbf{Assumption~1} is a mild requirement that is easily satisfied in near optimal region. Different from standard smoothness in optimization, it is trivial to justify \textbf{Assumption~2} by relating it to a quadratic loss function.

	\textbf{Lemma 1.} \textit{Let \textbf{Assumption 1} and \textbf{Assumption 2} hold. If $\left \| \theta - \theta^* \right \| \leq \epsilon$, then $\left \|\nabla_\theta \mathbb{E}_{(x,y)\sim \mathcal{P}}\left [ l\left ( f(\theta;x);y \right ) \right ] \right \| \leq L^2 \beta \epsilon$.
	}
	
	\textit{Proof.} This is a crucial result. So we sketch the proof as following. Using Jensen's inequality,
	\begin{equation*}
	\begin{split}
	& \left \|\nabla_\theta \mathbb{E}_{(x,y)\sim \mathcal{P}}\left [ l\left ( f(\theta;x);y \right ) \right ] \right \|^2 \\& \leq \mathbb{E}_{(x,y)\sim \mathcal{P}}\left [ \left \|\nabla_\theta  l\left ( f(\theta;x);y \right )  \right \|^2 \right ] \\
	& \leq \mathbb{E}_{(x,y)\sim \mathcal{P}}\left [ \left \|\nabla_{p} l\left ( p;y \right ) \nabla_\theta f(\theta;x)  \right \|^2 \right ], \text{where}~p = f(\theta;x)\\
	& \leq \mathbb{E}_{(x,y)\sim \mathcal{P}}\left [ \underset{\text{Cauchy-Schwarz inequality}}{\underbrace{\left \|\nabla_{p} l\left ( p;y \right )\right \|^2 \left \|\nabla_\theta f(\theta;x)  \right \|^2}} \right ] \\
	& \leq L^2 \mathbb{E}_{(x,y)\sim \mathcal{P}}\left [ \left \|\nabla_{p} l\left ( p;y \right )\right \|^2  \right ]
	\end{split}
	\end{equation*}
	
	Let $p=f(\theta;x)$ and $q=f(\theta^*;y)$. Using $\beta$-smoothness and $L$-Lipschitz property, we get
	\begin{equation*}
	\begin{split}
	\left \|\nabla_{p} l\left ( p;y \right )\right \| - \left \|\nabla_{q} l\left ( q;y \right )\right \|& \leq \left \|\nabla_{p} l\left ( p;y \right ) - \nabla_{q} l\left ( q;y \right )\right \| \\&\leq \beta \left \| p-q \right \| \\& \leq \beta L \left \| \theta - \theta^* \right \|.
	\end{split}
	\end{equation*}
	Since $\left \| \theta - \theta^* \right \| \leq \epsilon$, 
	\begin{equation*}
	\begin{split}
	\left \|\nabla_\theta \mathbb{E}_{(x,y)\sim \mathcal{P}}\left [ l\left ( f(\theta;x);y \right ) \right ] \right \|^2 \leq L^2 \mathbb{E}_{(x,y)\sim \mathcal{P}}\left [ \left ( \left \|\nabla_{q} l\left ( q;y \right )\right \| + L\beta \epsilon \right )^2  \right ].
	\end{split}
	\end{equation*}
	Upon substituting optimality condition, i.e., $\left \|\nabla_{q} l\left ( q;y \right )\right \|=0$, the above expression simplifies to
	\begin{equation*}
	\begin{split}
	\left \|\nabla_\theta \mathbb{E}_{(x,y)\sim \mathcal{P}}\left [ l\left ( f(\theta;x);y \right ) \right ] \right \| \leq L^2\beta \epsilon.
	\end{split}
	\end{equation*}
	This completes the proof of the theorem. \hfill $\square$
	
	\textbf{Lemma 1} provides an upper bound on the expected gradient over empirical distribution $\mathcal{P}$ in near optimal region. As the intermediate iterates ($\theta$) move closer to the optima ($\theta^*$), i.e., $\epsilon \rightarrow 0$, the gradient norm vanishes in expectation. This essentially resonates with the intuitive understanding of gradient descent. From another perspective, the issue of gradient descent inherently resides in near optimal region. We therefore ask a fundamental question: can we attain faster convergence without having to loose any empirical risk benefits? The following sections are intended to shed some light in this direction. 
	
	\textbf{Lemma 2.} \textit{Suppose \textbf{Assumption 1} holds. For a differentiable discriminator  $g(\psi;y)$, if  $\left \| g-g^* \right \| \leq \delta$, where $g^*\triangleq g(\psi^*)$ denote optimal discriminator, then $\left \|-\nabla_\theta \mathbb{E}_{x\sim \mathcal{P}_X}\left [ g\left ( \psi; f \left (\theta; x \right ) \right) \right ]\right \| \leq L \delta$.
	}
	
	\textit{Proof.} Using similar arguments from \textbf{Lemma 1},
	\begin{equation*}
	\begin{split}
	& \left \|-\nabla_\theta \mathbb{E}_{x\sim \mathcal{P}_X}\left [ g\left (\psi; f \left (\theta; x \right ) \right ) \right ]\right \|^2\\  & \leq \mathbb{E}_{x\sim \mathcal{P}_X}\left [ \left \|\nabla_\theta g\left (\psi; f \left (\theta; x \right ) \right )  \right \|^2 \right ]\\
	& \leq \mathbb{E}_{x\sim \mathcal{P}_X}\left [ \left \|\nabla_{p} g\left (\psi; p \right ) \right \|^2 \left \|\nabla_{\theta}f\left (\theta;x \right) \right \|^2 \right ], \text{ where } p=f\left(\theta;x\right)\\
	& \leq L^2 \mathbb{E}_{x\sim \mathcal{P}_X}\left [ \left \|\nabla_{p} g\left (\psi; p \right ) \right \|^2\right ] \\
	& \leq L^2 \mathbb{E}_{x\sim \mathcal{P}_X}\left [ \left (\left \|\nabla_{p}   g\left (\psi^*; p \right )\right \| + \delta\right )^2\right ]  \leq L^2 \delta^2
	\end{split}
	\end{equation*}
	Taking square root, $\left \|-\nabla_\theta \mathbb{E}_{x\sim \mathcal{P}_X}\left [ g\left (\psi; f \left (\theta; x \right ) \right ) \right ]\right \| \leq L \delta $, which finishes the proof. \hfill $\square$
	
	\textbf{Lemma 2} indicates that the expected gradient of purely adversarial generator does not produce erroneous gradients in the near optimal region, suggesting well behaved composite empirical risk~\cite{xue2018segan}.
	
	\textbf{Theorem 1.} \textit{Let us suppose \textbf{Assumption 1} and \textbf{Assumption 2} hold. If $\left \| \theta - \theta^* \right \| \leq \epsilon$ and $\left \| g-g^* \right \| \leq \delta$, then $\left \| \nabla_\theta  \mathbb{E}_{(x,y)\sim \mathcal{P}}\left [ l\left ( f(\theta;x);y \right ) -  g\left (\psi; f\left (\theta; x \right ) \right ) \right ]\right \| \leq \left ( L^2 \beta \epsilon + L \delta \right )$.
	}
	
	\textit{Proof.} By applying triangle inequality after simplification,
	\begin{equation*}
	\begin{split}
	& \left \| \nabla_\theta  \mathbb{E}_{(x,y)\sim \mathcal{P}}\left [ l\left ( f(\theta;x);y \right ) -  g\left (\psi; f\left (\theta; x \right ) \right ) \right ]\right \| \\& \leq \left \| \nabla_\theta  \mathbb{E}_{(x,y)\sim \mathcal{P}}\left [ l\left ( f(\theta;x);y \right )\right ] \right \| \\& \hspace{1cm}+ \left \| - \nabla_\theta \mathbb{E}_{(x,y)\sim \mathcal{P}}\left [ \left (\psi; f\left (\theta; x \right ) \right ) \right ] \right \| \\
	& \leq L^2 \beta \epsilon  + L \delta~(\text{\textbf{Lemma 1} and \textbf{Lemma 2}}),
	\end{split}
	\end{equation*} 
	which completes the statement of the theorem. \hfill$\square$
	
	To focus more on the empirical success of adversarial regularization, we study a simple convex-concave minimax optimization problem. It will certainly be interesting to borrow some ideas from the vast minimax optimization literature in various other settings~\cite{lin2019gradient,lin2020near,jin2019local,mertikopoulos2018cycles}. According to \textbf{Theorem 1}, the expected gradient of augmented objective does not vanish in the near optimal region, i.e., $\left \| \Delta \theta\right \| \rightarrow L \delta$ as $\epsilon \rightarrow 0$. In the current setting, the estimated gradients of $l(\theta)$ and $-g(\theta)$ at any instant during the optimization process are positively correlated. Thus, the gradients of augmented objective is lower bounded by $\left \| \nabla_\theta  \mathbb{E}_{(x,y)\sim \mathcal{P}}\left [ l\left ( f(\theta;x);y \right ) -  g\left (\psi; f\left (\theta; x \right ) \right ) \right ]\right \| \geq \left \| \nabla_\theta  \mathbb{E}_{(x,y)\sim \mathcal{P}}\left [ l\left ( f(\theta;x);y \right ) \right ]\right \|$. The upper and lower bounds of the intermediate iterates justify non-vanishing gradient in near optimal region. Having proven the contribution of discriminator in mitigating vanishing gradient, it seems natural to wonder whether adversarial regularization improves the iteration complexity.

	\subsection{Main Results: Asymptotic Iteration Complexity}
	\label{asym}
	In this section, we analyze global iteration complexity of sole supervision and adversarial regularization~\cite{zhang2019gradient,carmon2019lower}. The analysis is restricted to a deterministic setting. For a sequence of parameters $\left \{ \theta_k \right \}_{k\in \mathbb{N}}$, the complexity of a function $l(\theta)$ is defined as 
	
	\begin{equation*}
	\mathcal{T}_\epsilon\left ( \left \{ \theta_k \right \}_{k\in \mathbb{N}}, l \right ) \coloneqq \inf \left \{ k \in \mathbb{N} \mid \left \| \nabla l\left ( \theta_k \right ) \right \| \leq \epsilon \right \}. 
	\end{equation*}
	For a given initialization $\theta_0$, risk function $l$ and algorithm $A_\phi$, where $\phi$ denotes hyperparameters of training algorithm, such as learning rate and momentum coefficient, $A_\phi\left [ l ,\theta_0 \right ]$ denotes the sequence of iterates generated during training. We compute iteration complexity of an algorithm class parameterized by $p$ hyperparameters, $\mathcal{A} = \left \{ A_\phi  \right \}_{\phi \in \mathbb{R}^p}$ on a function class, $\mathscr{L}$ as 
	
	\begin{equation*}
	\begin{split}
	\mathcal{N}\left ( \mathcal{A}, \mathscr{L},\epsilon \right ) \coloneqq \inf_{A_\phi \in \mathcal{A}} \sup_{\theta_0 \in \left \{ \mathbb{R}^{h\times d_x}, \mathbb{R}^{d_y\times h} \right \},l\in \mathscr{L}}\mathcal{T}_\epsilon\left (A_\phi\left [ l ,\theta_0 \right ],l  \right ).
	\end{split}
	\end{equation*}
	We derive the asymptotic bounds under a less restrictive setting as introduced in~\cite{zhang2019gradient}. The new condition is weaker than commonly used Lipschitz smoothness assumption. Under this condition, the authors of~\cite{zhang2019gradient} aim to resolve the mystery of why adaptive gradient methods converge faster. We use this theoretical tool to study the asymptotic convergence of sole supervision and adversarial regularization in near optimal region. To circumvent the tractability issues in non-convex optimization, we follow the common practice of seeking an $\epsilon$-stationary point, i.e., $\left \| \nabla l\left ( \theta \right ) \right \| < \epsilon$.	We start by analyzing the iteration complexity of gradient descent with fixed step size. In this regard, we build on the assumptions made in \cite{zhang2019gradient}. To put more succinctly, let us recall the assumptions.\\
	\textbf{Assumption 3.} The loss $l$ is lower bounded by $l^* > -\infty$.\\
	\textbf{Assumption 4.} The function is twice differentiable. \\
	\textbf{Assumption 5.} ($(L_0,L_1)$-Smoothness). The function is $(L_0,L_1)$-smooth, i.e., there exist positive constants $L_0$ and $L_1$ such that $\left \| \nabla^2l\left ( \theta \right ) \right \| \leq L_0 + L_1 \left \| \nabla l\left ( \theta \right ) \right \|$.

	\textbf{Theorem 2.} \textit{Suppose the functions in $\mathscr{L}$ satisfy \textbf{Assumption 3, 4} and \textbf{5}. Given $\epsilon > 0$, the iteration complexity in sole supervision is upper bounded by $
		\mathcal{O}\left ( \frac{\left ( l(\theta_0)- l^* \right ) \left ( L_0+L_1 L^2 \beta \epsilon \right )}{\epsilon^2} \right ).
		$
	}
	
	\textit{Proof.} Refer to  Appendix~\ref{pr_th2}.
	
	\textbf{Corollary 1.} Using first order Taylor series, the upper bound in \textbf{Theorem 2} becomes $
	\mathcal{O}\left ( \frac{ l(\theta_0)- l^*}{h \epsilon^2} \right ).
	$
	
	\textit{Proof.} Refer to Appendix~\ref{pr_co1}.
	
	\textbf{Assumption 6.} (Existence of useful gradients) \textit{For arbitrarily small $\zeta > 0$, the norm of the gradients provided by discriminator is lower bounded by $\zeta$, i.e., $\left \| \nabla g \left ( \psi; f\left (\theta; x \right ) \right )\right \| \geq \zeta.$}
	
	\textbf{Assumption 6} requires the discriminator to provide useful gradients until convergence. It is a valid assumption in convex-concave minimax optimization problems. Also, it is trivial to prove this in the inner maximization loop under concave setting. In other words, the stated assumptions are mild, and derived from prior analyses for the sole purspose of maintaining consistency with existing literature. Keeping this in mind, we analyze the global iteration complexity in adversarial setting.
	
	\textbf{Theorem 3.} \textit{Suppose the functions in $\mathscr{L}$ satisfy \textbf{Assumption 3, 4 } and \textbf{5}. Given \textbf{Assumption 6} holds, $\epsilon > 0$ and $\delta \leq \frac{\sqrt{2\epsilon \zeta}}{L}$, the iteration complexity in adversarial regularization is upper bounded by $
		\mathcal{O}\left ( \frac{\left ( l(\theta_0)- l^* \right ) \left ( L_0+L_1 L^2 \beta \epsilon  \right )}{\epsilon^2 + 2\epsilon \zeta - L^2 \delta^2 } \right ).
		$
	}
	
	\textit{Proof.} Refer to Appendix~\ref{pr_th3}.
	
	\textbf{Corollary 2.} Using first order Taylor series, the upper bound in \textbf{Theorem 3} becomes $
	\mathcal{O}\left ( \frac{l(\theta_0)- l^*}{h \epsilon^2 + h \zeta \epsilon} \right ).
	$
	
	\textit{Proof.} Refer to Appendix~\ref{pr_co2}.	Since $2\epsilon \zeta - L^2 \delta^2  \geq 0$, the supervised learning with adversarial regularization has a \textit{tighter} global iteration complexity compared to sole supervision. In a simplified setup, one can easily verify this hypothesis by using first order Taylor's approximation as given by \textbf{Corollary~1}~and~\textbf{2}. In this case, $h \zeta \epsilon > 0$ ensures \textit{tighter} iteration complexity bound. This result is significant because it improves the convergence rates from $\mathcal{O}\left( \frac{1}{\epsilon^2}\right)$ to $\mathcal{O}\left( \frac{1}{\epsilon^2 + \epsilon \zeta}\right)$. Notice that for a too strong discriminator, \textbf{Assumption~6} does not hold. For a too weak discriminator, $\left \| g-g^* \right \|\leq \delta$ does not hold when $\delta$ is arbitrarily small. In these cases, the generator does not receive useful gradients from the discriminator to undergo accelerated training. However, for a sufficiently trained discriminator, i.e., $\left \| g-g^* \right \|\leq \delta \leq\frac{\sqrt{2\epsilon \zeta}}{L}$, the adversarial acceleration is guaranteed. Notably, the empirical risk and iteration complexity benefit from this provided the discriminator and the generator are trained alternatively as typically followed in practice.
	
	\subsection{Main Results: Sub-Optimality Gap}
	\label{flow}
	Here, we analyze the continuous time gradient flow in both approaches. The sub-optimality gap of generator and discriminator are defined by $\kappa(t) = \kappa(\theta(t)) \coloneqq l\left ( \theta(t)\right ) - l\left (  \theta^* \right ) $ and $\pi(t) = \pi(\theta(t)) \coloneqq g\left ( \theta^* \right ) - g\left (  \theta(t) \right ) $, respectively. In adversarial setting, $l(.)$ is a convex downward and $g(.)$ is a convex upward function. For clarity, we first analyze the gradient flow in sole supervision using common theoretic tools and then extend this analysis to adversarial regularization. 
	
	\textbf{Theorem 4.} \textit{In purely supervised learning, the sub-optimality gap at the average over all iterates in a trajectory of $T$ time steps is upper bounded by $
		\mathcal{O}\left ( \frac{\left \| \theta(0) - \theta^* \right \|^2}{2T} \right ).$
	}
	\textit{Proof.} Refer to Appendix~\ref{pr_th4}.
	
	\textbf{Theorem 5.} \textit{In supervised learning with adversarial regularization, the sub-optimality gap at the average over all iterates in a trajectory of $T$ time steps is upper bounded by
		\begin{equation*}
		\mathcal{O}\left ( \frac{\left \| \theta(0) - \theta^* \right \|^2}{2T} - \pi \left ( \frac{1}{T}\int_{0}^{T} \theta(t) dt \right ) \right ).
		\end{equation*}
	}
	\textit{Proof.} Refer to Appendix~\ref{pr_th5}.
	
	According to \textbf{Theorem 4} and \textbf{5}, the distance to optimal solution decreases rapidly in augmented objective when compared with purely supervised objective. Since sub-optimality gap is a non-negative quantity and $\pi \left ( \frac{1}{T}\int_{0}^{T} \theta(t) dt \right ) \geq 0 $, adversarial regularization has a \textit{tighter} sub-optimality gap. The tightness is controlled by the sub-optimality gap of adversary, $\pi(.)$ at the average over all iterates in the same trajectory. It is worth mentioning that the sub-optimality gap in adversarial regularization is at least as good as sole supervision which justifies the emprical gain in practice. Also, these theorems do not require all iterates to be within the tiny landscape of optimal empirical risk. The genericness of these theorems provides further evidence of empirical risk benefits in adversarial regularization.

	\subsection{Main Results: Provable Convergence}
	\label{conv_adv}
	This section covers the convergence guarantee of the minimax adversarial training under strongly-convex-strongly-concave and smooth nonconvex-nonconcave criteria. In this regard, we assume finite $\alpha$-moment of estimated stochastic gradients as the unbounded variance has a profound impact on optimization process~\cite{lacoste2012simpler}. At each iteration $k=1,\dots,T$, we denote unbiased stochastic gradient by $\mathfrak{g}_k = \mathfrak{g}(\theta_k) \coloneqq \nabla l(\theta_k,\xi) - \nabla g(\theta_k,\xi)$, where $\xi$ represents stochasticity. Here, we analyze rates for global clipping. One may wish to analyze this for coordinate-wise clipping~\cite{zhang2019adam}.
	
	\textbf{Assumption 7.} (Existence of $\alpha$-moment) Suppose we have access to gradients at each iteration. There exist positive real numbers $\alpha \in (1,2]$ and $G>0$, such that $\mathbb{E}\left [ \left \| \mathfrak{g(\theta)} \right \|^\alpha \right ]\leq G^\alpha$ for all $\theta$.
	
	\textbf{Theorem 6.} (Strongly-convex-strongly-concave convergence) \textit{Suppose \textbf{Assumption 7} holds. Let $\mathfrak{l} \left( \theta_k \right ) \triangleq l\left ( \theta_k \right ) - g\left ( \theta_k \right )$ is a $\mu$-strongly convex function. Let $\left \{ \theta_k \right \}$ be the sequence of iterates obtained using global clipping on SGD with zero momentum. Define the output to be $k$-weighted combination of iterates: $\bar{\theta} = \frac{\sum_{k=1}^{T} k \theta_{k-1}}{\sum_{k=1}^{T}k}$. If adaptive clipping $\tau_k = G k^{\frac{1}{\alpha}} \mu^{\frac{1}{\alpha}}$ and step size $\eta_k = \frac{5}{2\mu\left ( k+1 \right )} $, then the output iterate $\bar{\theta}$ satisfies 
		\begin{equation*}
		\begin{split}
		\mathbb{E}\left [ l\left ( \bar{\theta} \right ) \right ] - l\left ( \theta^* \right ) \leq \mathcal{O}\left ( G^2 \left ( \mu\left ( T+1 \right ) \right )^{\frac{2-2\alpha}{\alpha}} - \left ( g\left ( \theta^* \right ) - \mathbb{E}\left [ g\left ( \bar{\theta} \right ) \right ] \right ) \right).
		\end{split}
		\end{equation*}
	}
	\textit{Proof.} Refer to Appendix~\ref{pr_th6}.
	
	Observe that by eliminating adversary and setting $\alpha=2$, we recover exactly the SGD rate, i.e., $\mathcal{O}\left( \frac{G^2}{\mu T} \right)$~\cite{lacoste2012simpler}. Thus, adversarial regularization converges in strongly-convex-strongly-concave setting. It is determined by the convergence of the inner maximization loop in minimax optimization.

	\textbf{Theorem 7.} (Nonconvex-nonconcave convergence) \textit{Suppose \textbf{Assumption 3.1} and \textbf{3.2} hold. Let $\mathfrak{l} \left( \theta_k \right ) \triangleq l\left ( \theta_k \right ) - g\left ( \theta_k \right )$ is a possible $L$-smooth function and $\{ \theta_k\}$ be the sequence of iterates obtained using global clipping on SGD with zero momentum. Given constant clipping $\tau_k = G\left ( \eta_k L  \right )^{\frac{-1}{\alpha}}$ and constant step size $\eta_k = \left ( \frac{R^{\alpha}_{0}L^{2-2\alpha}}{G^2 T^\alpha} \right )^{\frac{1}{3\alpha -2}}$, where $R_0 = l(\theta_0) - l(\theta^*)$, the sequence $\{ \theta_k\}$ satisfies
		\begin{equation*}
		\begin{split}
		\frac{1}{T}\sum_{k=1}^{T}\mathbb{E}\left [ \left \| \nabla l\left ( \theta_{k-1} \right ) \right \|^2 \right ] \leq \mathcal{O}\left (G^{\frac{2\alpha}{3\alpha-2}} \left ( \frac{R_0 L}{T} \right )^{\frac{2\alpha-2}{3\alpha-2}} - \frac{1}{T} \sum_{k=1}^{T} \mathbb{E}\left [ \left \| \nabla g(\theta_{k-1}) \right\|^2  \right ]  \right ).
		\end{split}
		\end{equation*}
	}
	\textit{Proof.} Refer to Appendix~\ref{pr_th7}.
	
	By setting $\alpha = 2$ and discarding adversarial acceleration, we obtain the standard SGD rate, $\mathcal{O}\left( \frac{G}{\sqrt{T}} \right)$. It is important to heed the fact that adversarial regularization converges under nonconvex-nonconcave criterion as well. To this end, we have established that augmented objective is \textit{guaranteed} to converge under strongly-convex-strongly-concave and nonconvex-noncave criteria provided the assumptions are satisfied. These guarantees provide more insights to our understanding of adversarial training in practice. While this paper studies minimax optimization under nonconvex-smooth settings, it will be interesting to derive convergence guarantees under nonconvex-nonsmooth setting.

	\subsection{Main Results: Generalization Error}
	\label{gap}
	Motivated by the role of over-parametrization in generalization~\cite{neyshabur2017exploring,nagarajan2017generalization,neyshabur2019role}, we study the generalization behavior of adversarial regularization. We use Rademacher complexity to get a bound on generalization error. Since it depends on hypothesis class, we use a set of restricted parameters of trained networks to get a tighter bound on generalization. The restricted set of parameters is defined as
	
	\begin{equation*}
	\begin{split}
	\mathcal{W} = \left \{ \left ( V,U \right ) \vert V \in \mathbb{R}^{d_y \times h}, U\in \mathbb{R}^{h\times d_x}, \left \| v_i \right \|\leq \alpha_i, \left \| u_i -u_i^0\right \|\leq \beta_i \right \},
	\end{split}
	\end{equation*} 
	where $i = 1,2,\dots,h$. Here, $v_i\in \mathbb{R}^{dy}$ and $u_i \in \mathbb{R}^{dx}$ denote vector representation of each neuron in the top layer and hidden layer, respectively. Thus, the restricted hypothesis class becomes
	
	\begin{equation*}
	\begin{split}
	\mathcal{F}_\mathcal{W} = \left \{ V[Ux]_+ \vert \left ( V,U \right ) \in \mathcal{W} \right \},
	\end{split}
	\end{equation*}
	where $[.]_+$ represents ReLU activation. For any hypothesis class $\mathcal{F}$, let $l~o~\mathcal{F}$ denote the composition of loss function and hypothesis class. The following bound holds for any $f\in \mathcal{F}_{\mathcal{W}}$ over $m$ training samples with probability $1-\delta$.
	
	\begin{equation*}
	\begin{split}
	\mathbb{E}_{(x,y)\sim \mathcal{D}}\left [ l~o~f \right ] \leq \frac{1}{m}\sum_{i=1}^{m} l\left ( f(x);y \right ) + 2 \mathcal{R}_{\mathcal{S}}\left ( l~o~\mathcal{F}_{\mathcal{W}} \right ) + 3 \sqrt{\frac{ln(2/\delta)}{2m}},
	\end{split}
	\end{equation*}
	where $\mathcal{R}_{\mathcal{S}}(\mathcal{H})$ is the Rademacher complexity of a hypothesis class $\mathcal{H}$ with respect to training set $\mathcal{S}$.
	
	\begin{equation*}
	\begin{split}
	\mathcal{R}_{\mathcal{S}}\left ( \mathcal{H} \right )=\frac{1}{m} \mathbb{E}_{\xi_i \in \left \{ \pm 1 \right \}^m} \left [ \sup_{f\in \mathcal{H}} \sum_{i=1}^{m} \xi_i f(x_i)   \right ].
	\end{split}
	\end{equation*}
	
	\textbf{Relative Generalization Error:} We define relative generalization error as 
	
	\begin{equation*}
	\begin{split}
	e_{gen,r} = \left ( \mathbb{E}_{(x,y)\sim \mathcal{D}}\left [ l~o~f \right ] - \frac{1}{m}\sum_{i=1}^{m} l\left ( f(x);y \right )  \right ) \times N^*.
	\end{split}
	\end{equation*}
	To be consistent with \citeauthor{neyshabur2019role} while studying generalization, we assume $l(f(\theta;x);y)$ be a locally $K$-Lipschitz function, i.e., given $y \in Y$, $\left \|\nabla l(f(\theta;x);y)\right \| \leq K,~\forall~\theta$. Using $K$-Lipschitz property of loss function $l$ in \textbf{Lemma~9} of~\citeauthor{neyshabur2019role}, one can easily prove that the Rademacher complexity of $l~o~\mathcal{F}_{\mathcal{W}}$ is bounded as
	
	\begin{equation*}
	\begin{split}
	&\mathcal{R}_{\mathcal{S}}\left ( l~o~\mathcal{F}_{\mathcal{W}} \right ) \\& \leq \frac{2K\sqrt{d_y}}{m}\sum_{j=1}^{h} \alpha_j \left ( \beta_j \left \| X \right \|_F + \left \| u_j^0 X \right \|_2 \right ) \\ & \leq \frac{2K\sqrt{d_y}}{\sqrt{m}}\left \| \alpha  \right \|_2 \left ( \left \| \beta \right \|_2 \sqrt{\frac{1}{m} \sum_{i=1}^{m}\left \| x_i \right \|_2^2 } + \sqrt{\frac{1}{m} \sum_{i=1}^{m}\left \| U^0 x_i \right \|_2^2 } \right ). 
	\end{split}
	\end{equation*}
	Adapted to current setting, the generalization error becomes
	
	\begin{equation*}
	\begin{split}
	\mathcal{O}\left ( \left \| U^0 \right \|_2 \left \| V \right \|_F + \left \| U-U^0 \right \|_F \left \| V \right \|_F +\sqrt{h} \right ).
	\end{split}
	\label{comb_meas}
	\end{equation*}
	Next, we empirically verify the required assumptions and corresponding theoretical results. 
	
	\begin{figure}[t]
		\centering
		\includegraphics[width=0.97\columnwidth]{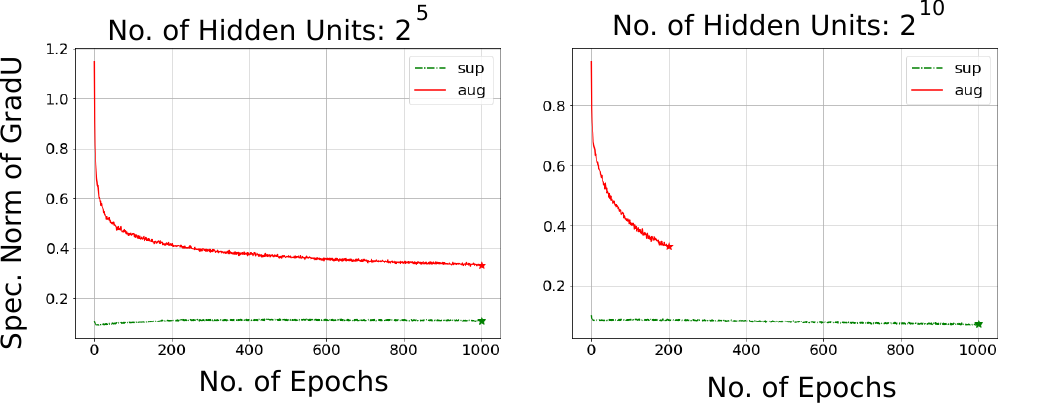}
		\caption{Comparison of gradients --- supervised (sup) and augmented (aug) --- in the \textit{hidden layer} on MNIST. Adversarial regularization mitigates vanishing gradient issue.}
		\label{gradu}
	\end{figure}
	
	\begin{figure}[t]
		\centering
		\includegraphics[width=0.96\columnwidth]{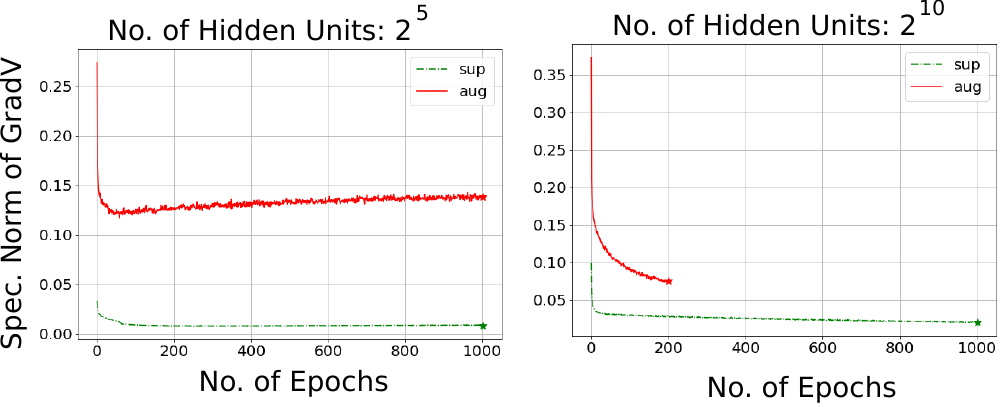}
		\caption{Comparison of gradients --- supervised (sup) and augmented (aug) --- in the \textit{top layer} on MNIST. Adversarial regularization mitigates vanishing gradient issue.}
		\label{gradv}
	\end{figure}
	
	\section{Experiments}
	Our experiments aim to answer the following questions\footnote{While the preliminary observations are reported in the main paper, additional experimental results are supplied in the appendix.
	}.
	\begin{itemize}
		\item How does adversarial regularization mitigate vanishing gradients in the near optimal region?
		\item How does adversarial regularization accelerate training?
		\item How does adversarial regularization achieve tighter sub-optimality gap?
		\item How does adversarial regularization converge under practical settings?
	\end{itemize}

	\subsection{Results on MNIST}
	Figure~\ref{gradu}~and~\ref{gradv} provide empirical evidence of the vanishing gradient issue, and how adversarial regularization helps circumvent this. In all the experimented architectures, the spectral norm of gradients estimated in purely supervised objective is smaller than adversarial learning. This is consistent with the theoretical analyses in Section~\ref{method}. The main reason for such non-vanishing gradient is the feedback signal from discriminator. Further, the rate of convergence is at least as good as sole supervision, as marked by $\star$ in Figure~\ref{gradu}~and~\ref{gradv}.

\begin{figure}[t]
		\centering
		\includegraphics[width=\columnwidth]{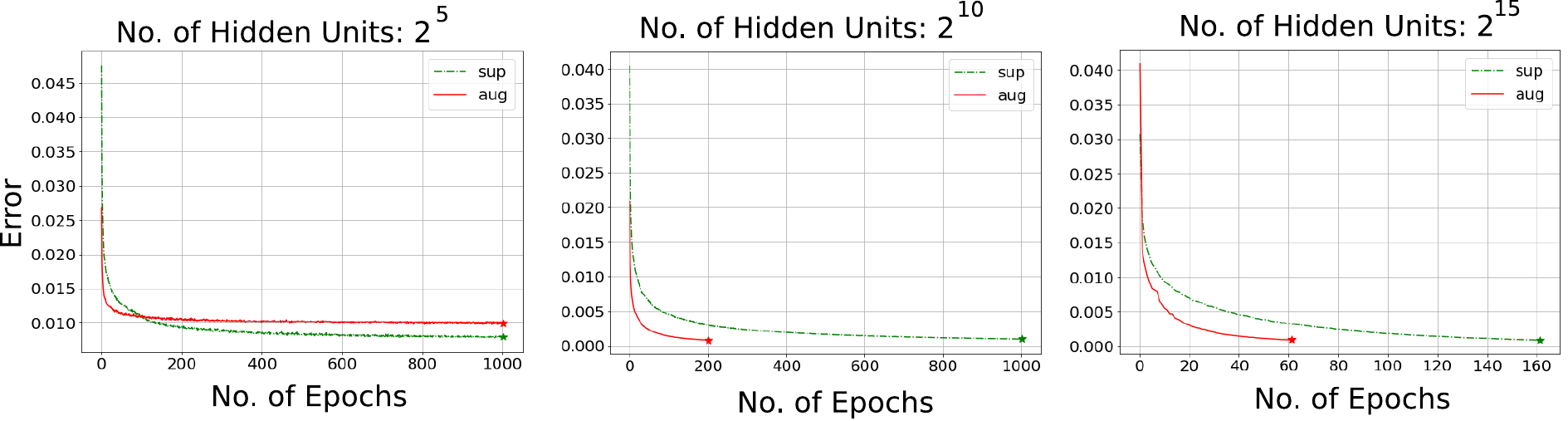}
		\caption{Comparison of optimal empirical risk on MNIST. Adversarial regularization converges faster.}
		\label{emp_risk}
	\end{figure}

	Figure~\ref{emp_risk} offers experimental support to better empirical risk in adversarial setting. Here, we observe the significance of near optimal region, i.e., $\epsilon$ with 32 hidden units. Since the expressive power of such a network is very small in both approaches, evidently neither of those meets the convergence criteria. However, as the capacity increases the supervised cost, which is common in both approaches, guides them to a tiny landscape around optimum and thereby, it satisfies the assumptions of \textbf{Theorem 1}. It is to be noted that the tightness of the reported bounds is asserted in the near optimal region. This is evident from the stability of the Lipschitz constant $L$ over iterations as shown in Figure~\ref{gradu} and \ref{gradv}. Under this circumstance, the optimal empirical risk in augmented objective can be provably better than sole supervision as predicted by the proposed theorems. Figure~\ref{emp_risk} supports this theory as augmented objective consistently achieves better performance either by risk or by rate of convergence for networks with sufficient expressive power.
	
\begin{figure}[t]
		\centering
		\includegraphics[width=\columnwidth]{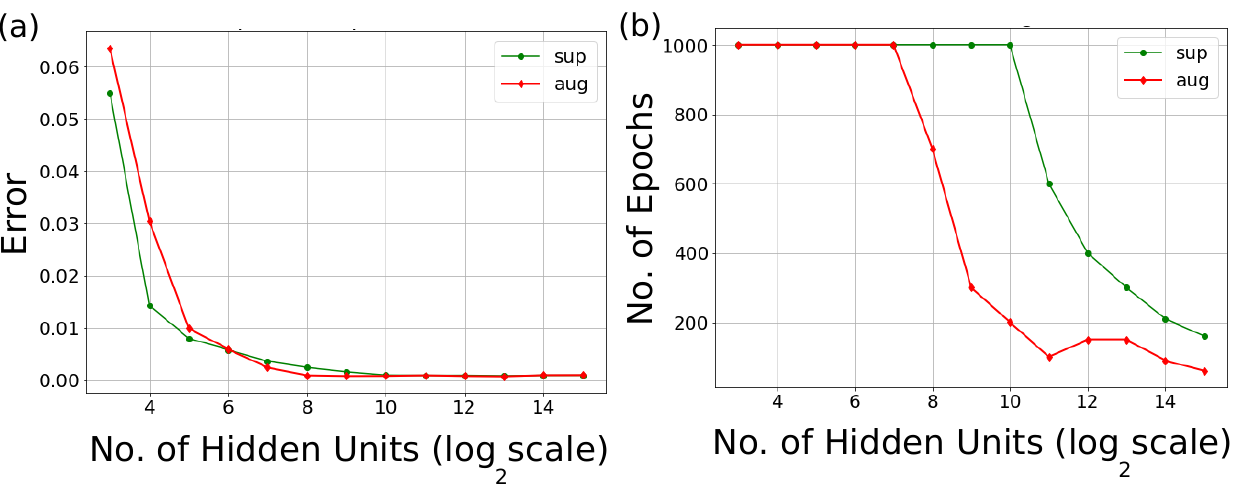}
		\caption{Comparison on MNIST. (a)~Optimal empirical risk. (b)~Iteration Complexity. Adversarial regularization attains tighter $\epsilon$-stationary point at an optimal rate.}
		\label{emp_conv}
	\end{figure}

	Furthermore, we compare the optimal empirical risk and iteration complexity with different number of hidden units in Figure~\ref{emp_conv}. To better interpret the theorems, one can infer from Figure~\ref{emp_conv}~(a) that the value of $\epsilon$ in \textbf{Theorem 1} is approximately equal to 0.005. The number of epochs required to find a first order stationary point in adversarial learning is always less than or equal to supervised learning, which validates our theorems. The value of $\epsilon$ is more relevant to the present body of analysis as it is a major part of the inverse mapping in practical scenarios. Moreover, it is not hard to estimate $\delta$ in some rare occurences where the mapping function is approximated by the discriminator.

	\subsection{Results on CIFAR10}
\begin{figure}[t]
		\centering
		\includegraphics[width=\columnwidth]{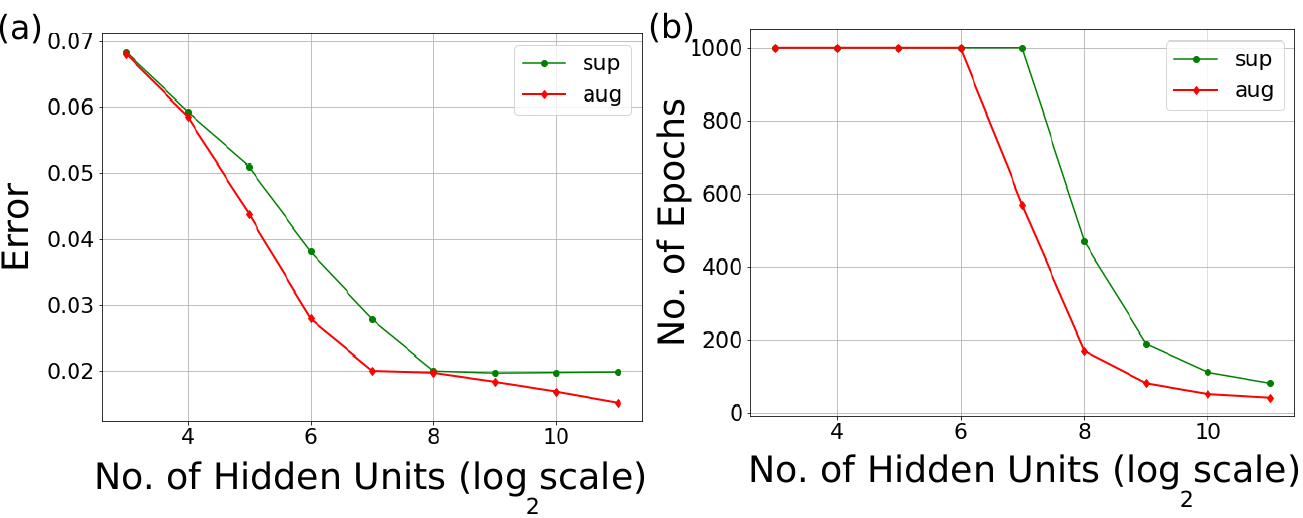}
		\caption{Comparison on CIFAR10. (a)~Optimal empirical risk. (b)~Iteration Complexity. Adversarial regularization attains tighter $\epsilon$-stationary point at an optimal rate.}
		\label{emp_conv_cifar}
	\end{figure}

	These theorems also justify the experiments conducted on CIFAR10 dataset. As shown in Figure~\ref{emp_conv_cifar}, supervised learning with adversarial regularization performs better than sole supervision both in terms of optimal empirical risk and  iteration complexity. Here, $\epsilon$ is approximately equal to 0.06.

	\subsection{Results on Generalization Error}
\begin{figure}[t]
		\centering
		\includegraphics[width=\columnwidth]{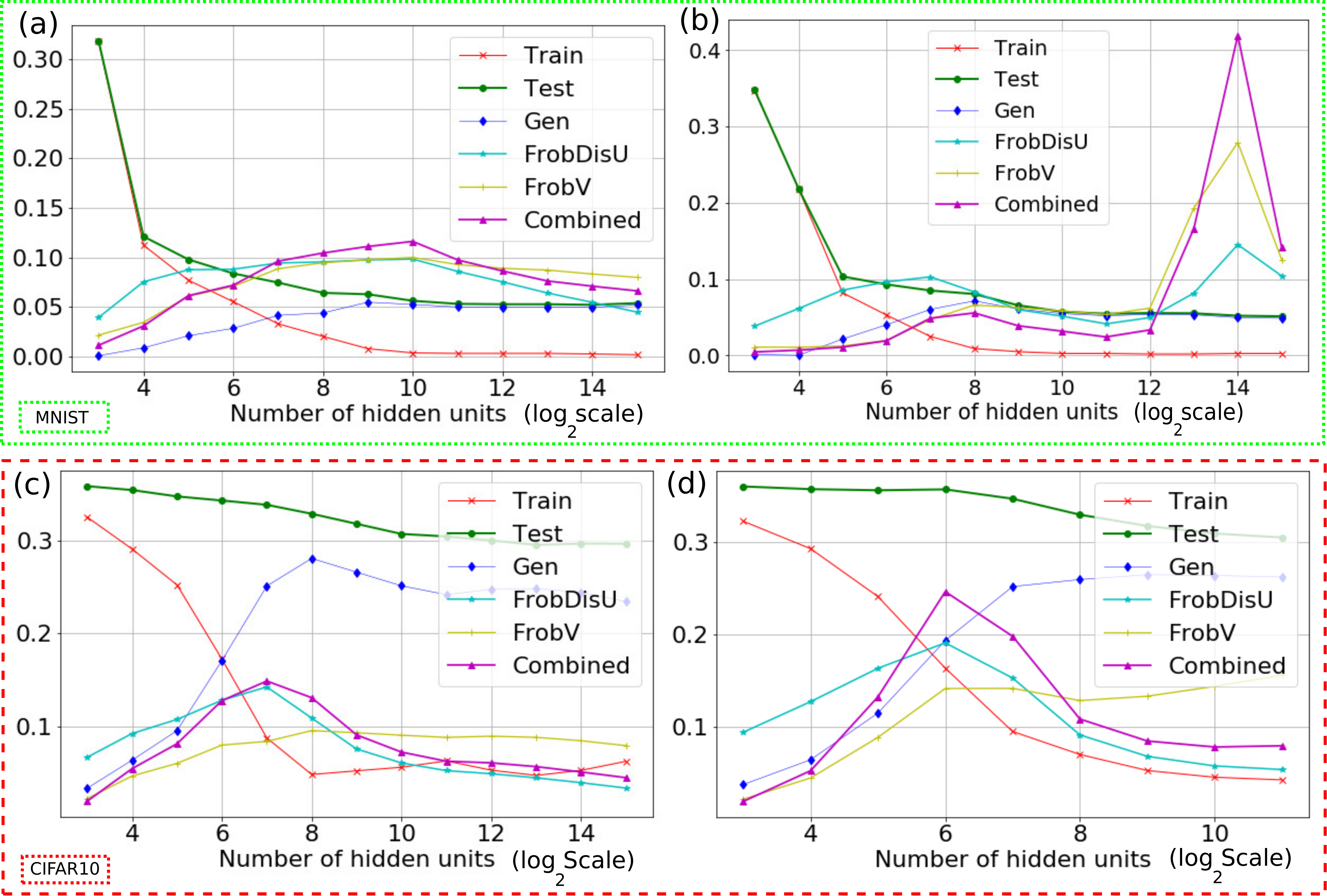}
		\caption{Generalization error on MNIST and CIFAR10. Adversarial training requires new generalization bound.}
		\label{sup_gen}
	\end{figure}

	The generalization trend in sole supervision is shown in Figure~\ref{sup_gen}(a)~and~\ref{sup_gen}(c). As per equation~(\ref{comb_meas}), the combined measure of Frobenius norm of top layer, i.e., $\left \| V \right \|_F$ and distance from initialization of hidden layer, i.e., $\left \| U-U^0 \right \|_F$ explains the generalization gap on MNIST and CIFAR10. We verify this measure in our experimental setting and study whether it can explain generalization in adversarial learning. Note that adversarial learning and sole supervision share exactly same mapping function ($f$), learning algorithm (SGD+momentum) and empirical data distribution ($S$). The generalization bound, therefore, is expected to explain the generalization error in adversarial learning with expert regularization. However, as shown in Figure~\ref{sup_gen}(b)~and~\ref{sup_gen}(d), this bound does not fully explain the generalization error observed in adversarial learning. 
	
	
\begin{figure}[t]
		\centering
		\includegraphics[width=\columnwidth]{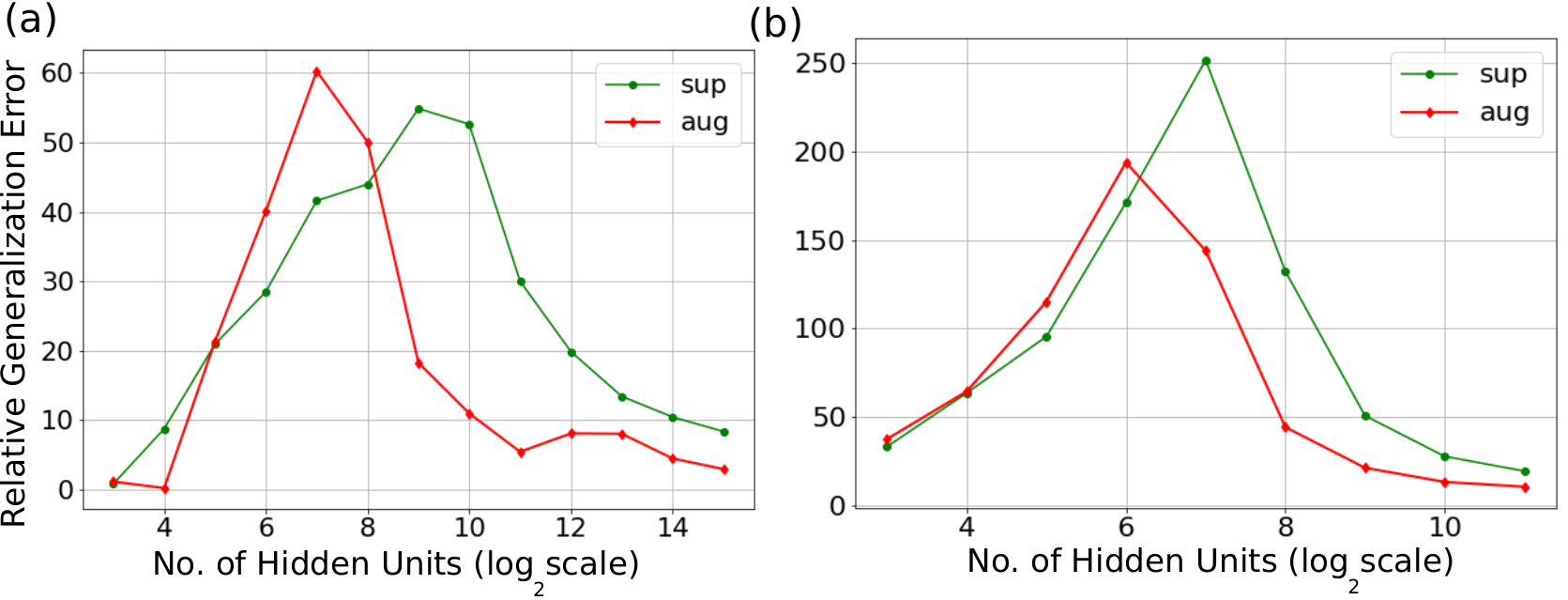}
		\caption{Relative generalization. (a) MNIST. (b) CIFAR10. Augmented objective has better relative generalization error.}
		\label{rel_gen}
	\end{figure}

	In Figure~\ref{rel_gen}, we observe that the relative generalization error of adversarial regularization can be better than sole supervision. This is feasible for a network with sufficient expressive power to achieve near optimal convergence.	
	
	\section{Discussion}
	In this study, we investigated the reason behind slow convergence of purely supervised learning in near optimal region, and how adversarial regularization circumvents this issue. Further, we explored several crucial properties at this juncture of understanding the role of adversarial regularization in supervised learning. Particularly intriguing was the genericness of these theorems around the central theme. To make a fair assessment, standard theoretic tools were employed in all the theorems. From theoretical perspective, the iteration complexity, gradient flow, provable convergence guarantee, and the analysis of generalization error provided further insights to the empirical findings of adversarial regularization as independently reported in previous works.
	
	While these theoretical analyses provided several key insights to better understand the practical success of adversarial regularization, it is far from being conclusive. Moreover, it paves the way for several open questions: (i) What explains the generalization behavior in adversarial learning? (ii)  Does adversarial regularization improve sample complexity? In this paper, we do not explain generalization gap and sample complexity. Nevertheless, it will be interesting to understand the effect of implicit gradient estimation by an adversary on these theoretic puzzles.

	\section{Broader Impact}
	In this paper, we primarily focused on understanding the role of adversarial regularization in supervised learning. At a fundamental level, we provided a theoretical justification supported by empirical evidence to corroborate commonly observed phenomena in practice. We believe this work does not present any foreseable societal consequence. 
	
	\bibliography{aaai21}
	
	\appendix
	\onecolumn
	\section*{Appendix}
	\section{More Experiments}
	\label{exps}
	This section contains additional results and discussion to support the theoretical findings on sole supervision and adversarial regularization.
	
	\subsection{Training Details}
	\label{impl}
	The majority of the experiments are conducted on two layer neural networks with ReLU activation function. For completeness however, we experiment with practical neural network architectures. We do not use weight decay, dropout or normalization in these networks. Experiments are conducted on MNIST and CIFAR10 datasets. We use SGD with momentum 0.9, batch size 64 and fixed learning rate of 0.01 for MNIST and CIFAR10.  The convergence criterion is set to be mean square error of 0.001 for MNIST and 0.02 for CIFAR10. We train on both datasets for a maximum of 1000 epochs, or until convergence. In these settings, 13 architectures with the number of hidden units ($h$) ranging from $2^3$ to $2^{15}$ are trained on both datasets. All parameters are initialized from uniform distribution. The experiments are conducted on a Linux system with 64GB RAM and 2 x V100 gpus using PyTorch library.

	\subsection{Experimental Results}
	\label{res}
	\subsubsection{Results on MNIST}
\begin{figure}[t]
		\centering
		\includegraphics[width=0.8\columnwidth]{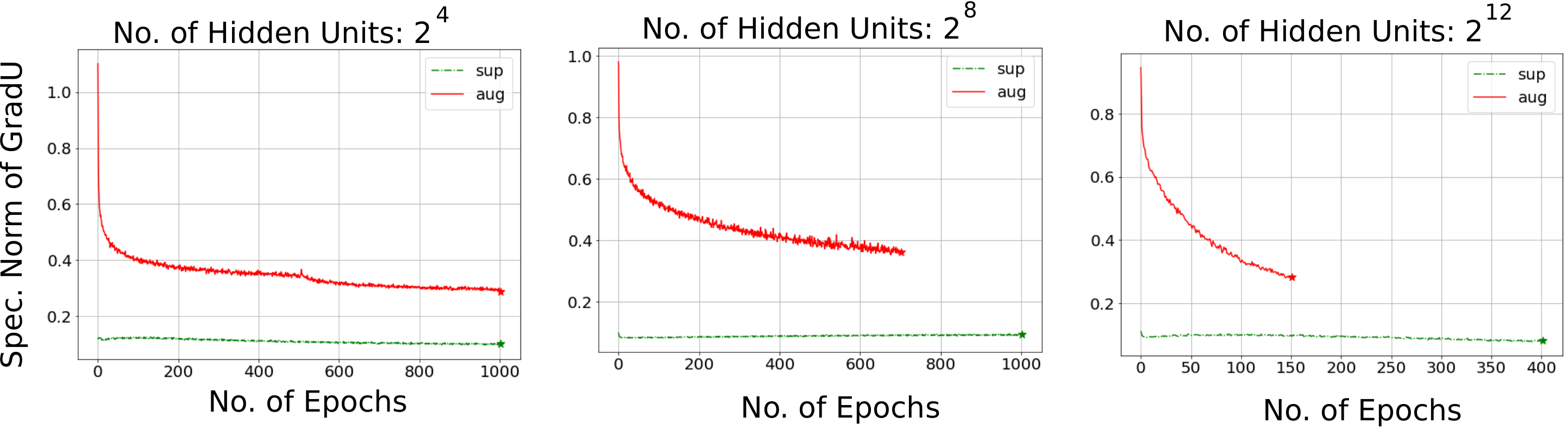}
		\caption{Comparison of gradient updates between supervised and augmented objective as observed in the \textit{hidden layer} on MNIST.}
		\label{mnist_gradu2}
	\end{figure}

\begin{figure}[t]
		\centering
		\includegraphics[width=0.8\columnwidth]{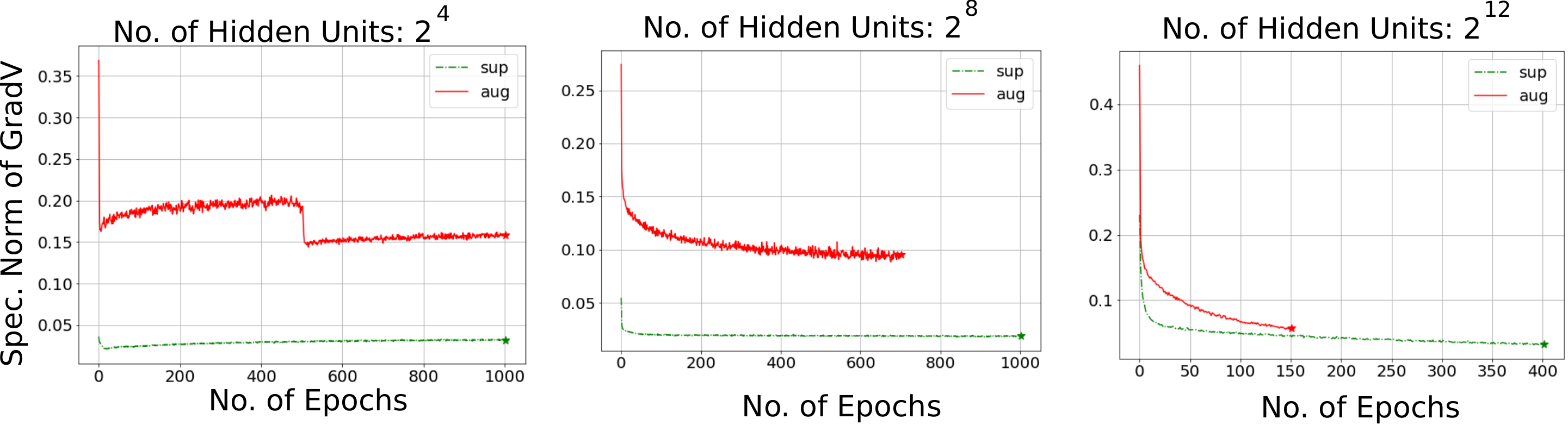}
		\caption{Comparison of gradient updates between supervised and augmented objective as observed in the  \textit{top layer} on MNIST.}
		\label{mnist_gradv2}
	\end{figure}

\begin{figure}[t]
		\centering
		\includegraphics[width=0.8\columnwidth]{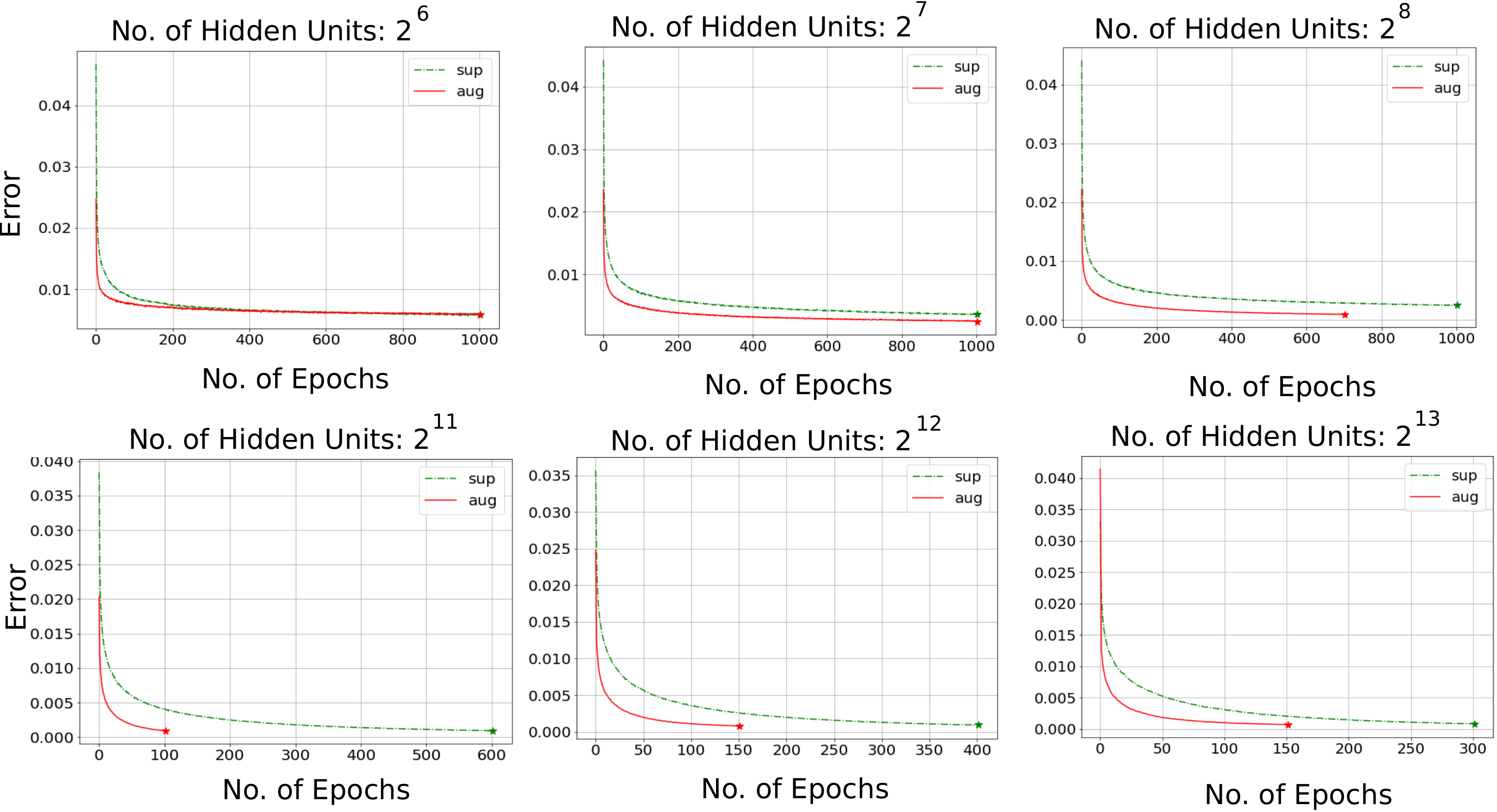}
		\caption{Comparison of optimal empirical risk on MNIST.}
		\label{mnist_emp_risk2}
	\end{figure}

	As shown in Figure~\ref{mnist_gradu2}~and~\ref{mnist_gradv2}, the estimated gradient in SGD+momentum vanishes within the tiny landscape of optimal empirical risk. Further, the adversarial regularization accelerates gradient updates and attains minimal empirical risk compared to sole supervision. It is evident from Figure~\ref{mnist_emp_risk2} where we observe this particular phenomenon across a wide variety of architectures. One may argue that the difference in empirical risk is minimal. However, it is always better to discover a first order stationary point relatively faster without having to loose any risk benefits. From another perspective, the notion of multiple critical points in deep neural networks acts in favor of adversarial learning that allows faster convergence. It seems to us as an interesting line of future work.

	\subsubsection{Results on CIFAR10}
\begin{figure}[!t]
		\centering
		\includegraphics[width=0.8\columnwidth]{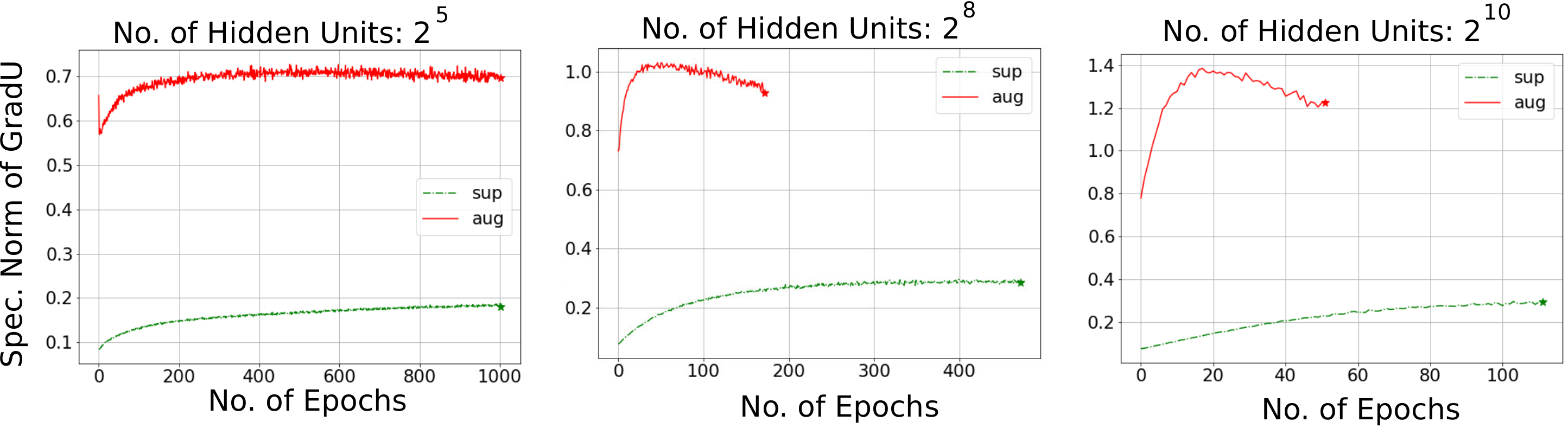}
		\caption{Comparison of gradient updates between supervised and augmented objective as observed in the \textit{hidden layer} on CIFAR10.}
		\label{cifar_gradu1}
	\end{figure}

\begin{figure}[!t]
		\centering
		\includegraphics[width=0.8\columnwidth]{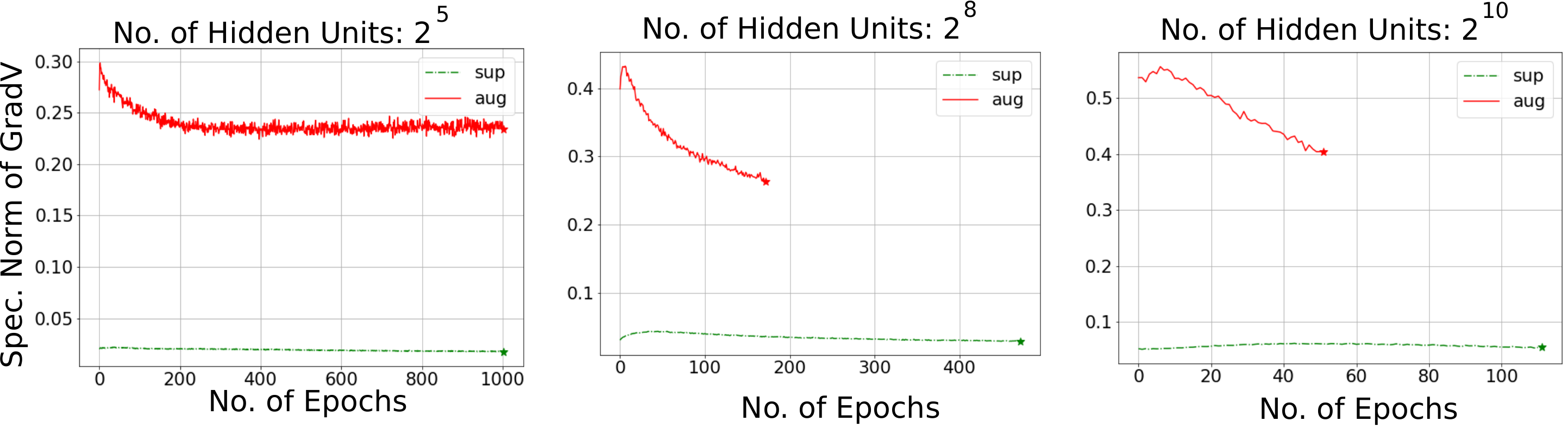}
		\caption{Comparison of gradient updates between supervised and augmented objective as observed in the  \textit{top layer} on CIFAR10.}
		\label{cifar_gradv1}
	\end{figure}

\begin{figure}[!t]
		\centering
		\includegraphics[width=0.8\columnwidth]{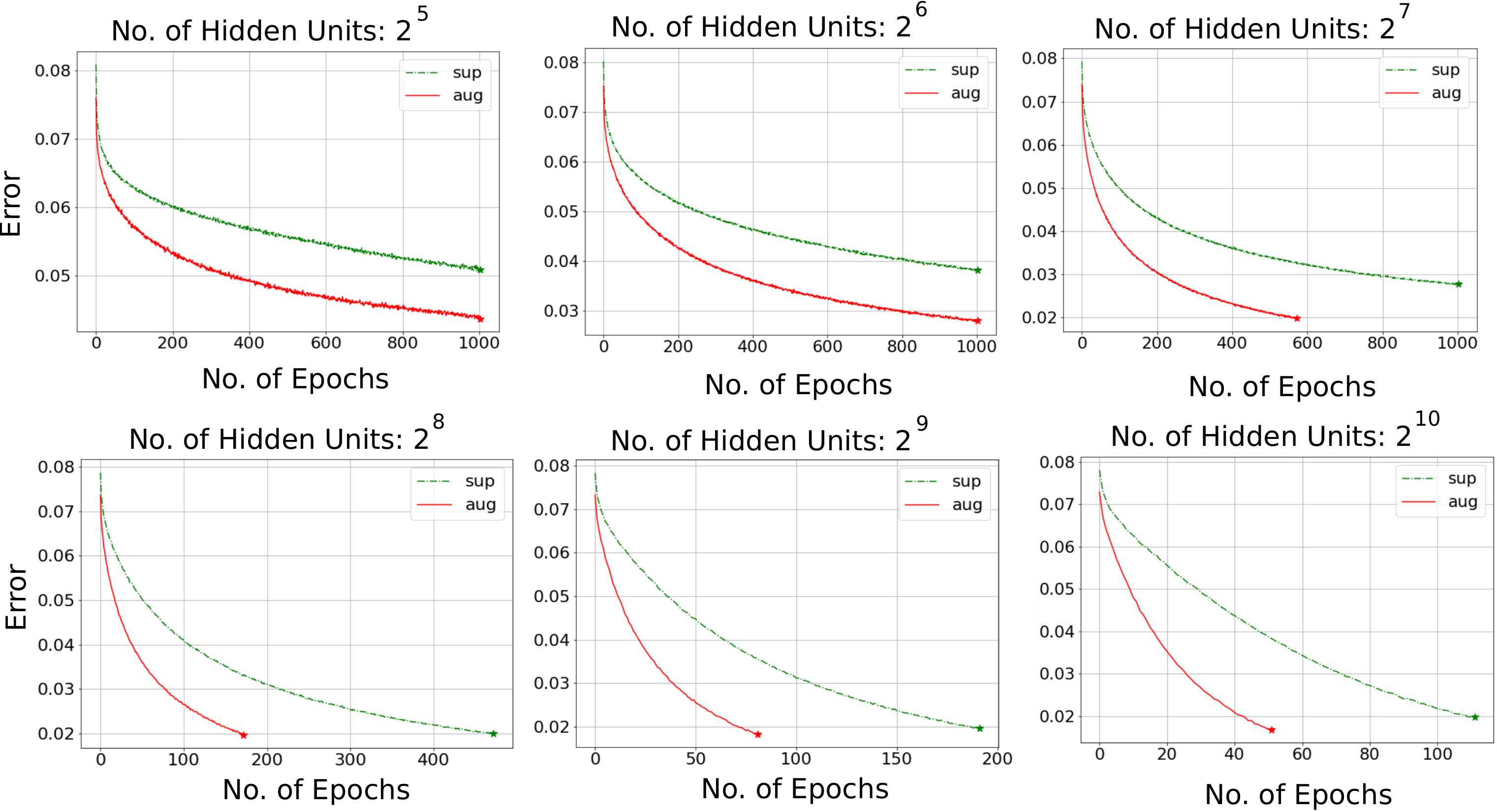}
		\caption{Comparison of optimal empirical risk on CIFAR10.}
		\label{cifar_emp_risk1}
	\end{figure}

	Similar to MNIST, we also observe vanishing gradient issue on CIFAR10, which is shown in Figure~\ref{cifar_gradu1}~and~\ref{cifar_gradv1}. Figure~\ref{cifar_emp_risk1} illustrates how model capacity correlates with empirical risk and thereby, satisfies the assumption of \textbf{Theorem~1}. Across a wide variety of architectures, observe that supervised learning with adversarial regularization can be better than sole supervision both in terms of optimal empirical risk and  iteration complexity as predicted by our theory. As shown in Figure~\ref{cifar_emp_risk1}, though both methods start with almost same initial empirical risk, augmented objective traverses through a shorter path and attains minimal risk upon convergence. The slight difference in error at the begining is mainly due to adversarial acceleration in the first step itself.
	
	\subsubsection{Results on Various Networks}
	\label{nets}
\begin{figure}[t]
		\centering
		\includegraphics[width=0.6\columnwidth]{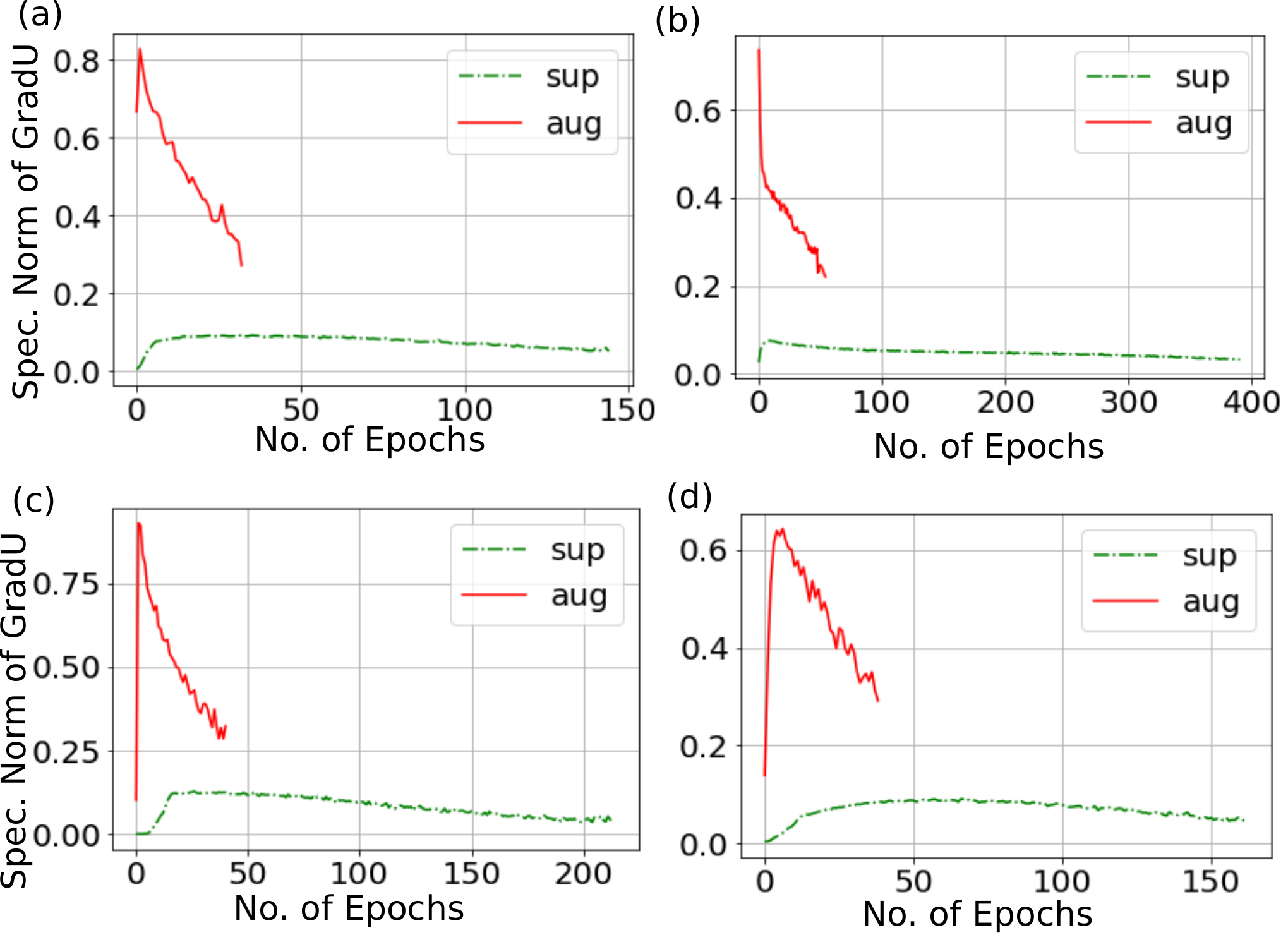}
		\caption{Comparison of gradient updates between supervised and augmented objective as observed in the \textit{first layer} on MNIST. (a) Multi-Layer Perceptron. (b) Exponential Activation. (c) Residual Network. (d) Dense Network.}
		\label{mnist_gradu3}
	\end{figure}

\begin{figure}[t]
		\centering
		\includegraphics[width=0.6\columnwidth]{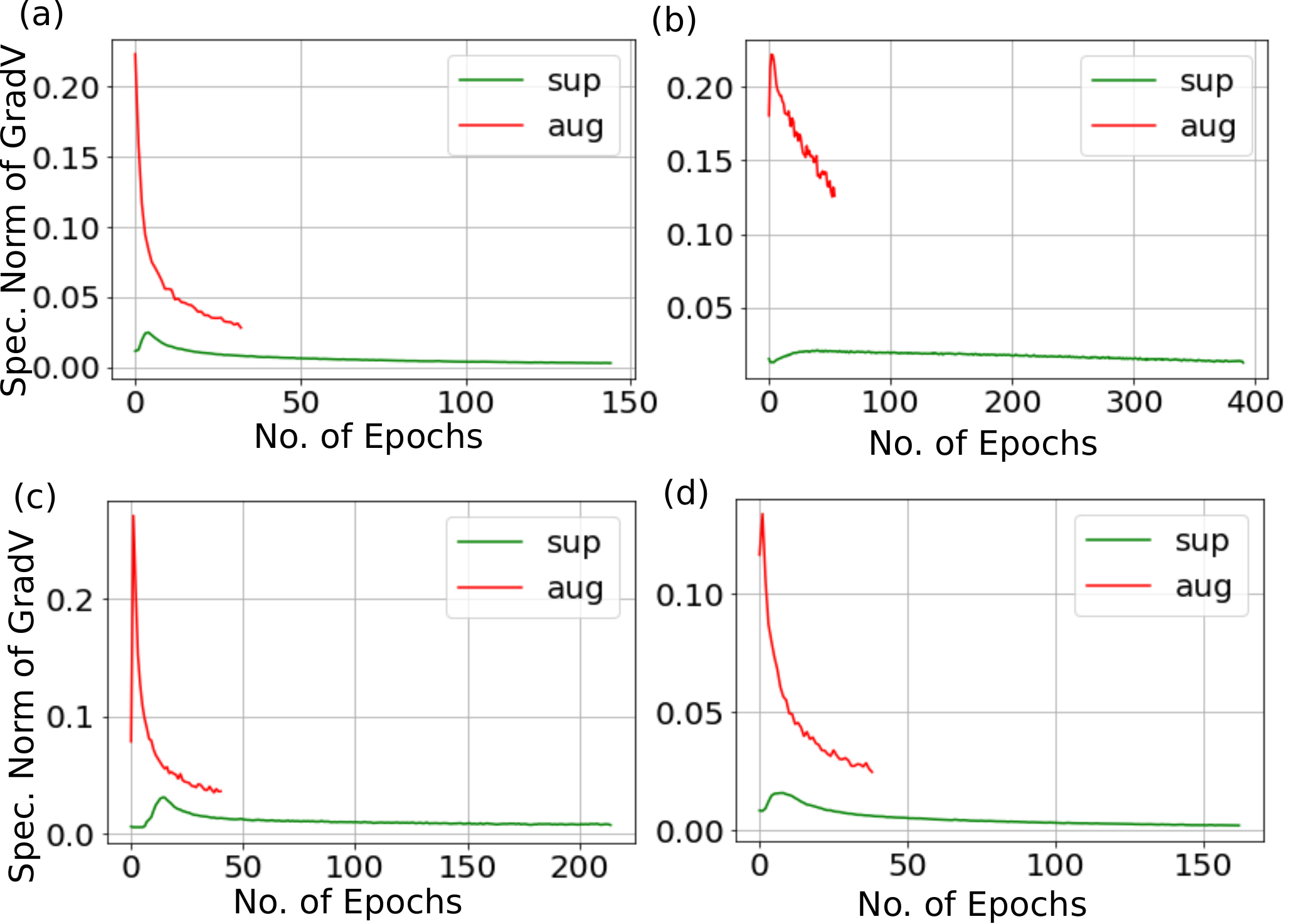}
		\caption{Comparison of gradient updates between supervised and augmented objective as observed in the \textit{last layer} on MNIST. (a) Multi-Layer Perceptron. (b) Exponential Activation. (c) Residual Network. (d) Dense Network.}
		\label{mnist_gradv3}
	\end{figure}

\begin{figure}[t]
		\centering
		\includegraphics[width=0.8\columnwidth]{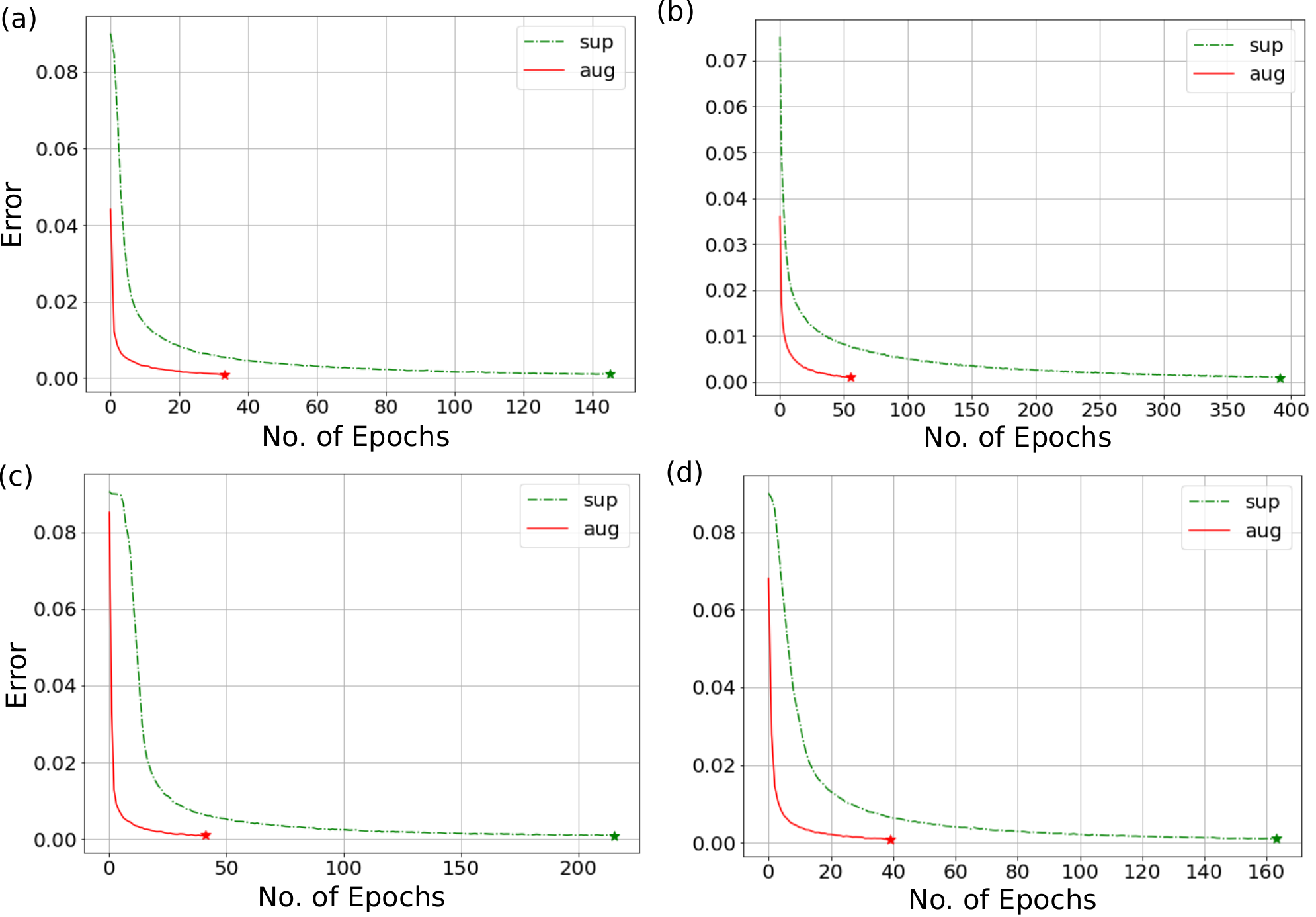}
		\caption{Comparison of optimal empirical risk on MNIST. (a) Multi-Layer Perceptron. (b) Exponential Activation. (c) Residual Network. (d) Dense Network.}
		\label{mnist_emp_risk3}
	\end{figure}

	To study the impact of these findings on more realistic scenarios, we experiment on various network configurations. As shown in Figure~\ref{mnist_gradu3}~and~\ref{mnist_gradv3}, the issue of vanishing gradient is persistent across these experimented configurations. Furthermore, the discussion on adversarial acceleration is also supported by Figure~\ref{mnist_emp_risk3}. In addition, Table~~\ref{prac_nets} shows that the proposed hypothesis: \textit{adversarial regularization achieves tighter $\epsilon$-stationary point at an optimal rate} holds under practical circumstances. More specifically, we observe accelerated gradient updates not only in two layer ReLU networks, but also in deep MLP with exponential linear activations, convolution layers, skip connections, dense connections, $L_1$ regularized networks, and $L_2$ regularized networks. Thus, augmented objective owes its performance benefits to adversarial learning at a fundamental level.
	
	\begin{table}[t]
		\caption{Hypothesis Testing on Various Network Configurations}
		\label{prac_nets}
		\resizebox{\textwidth}{!}{%
			\begin{tabular}{llllllll}
				\hline
				Architecture       & No. Layer & Activation & No. ResBlock & No. DenseBlock & \begin{tabular}[c]{@{}l@{}}No. Epoch\\       Sup\end{tabular} & \begin{tabular}[c]{@{}l@{}}No. Epoch\\      Aug\end{tabular} & Hypothesis   \\ \hline
				MLP-Deep           & 6         & ELU        & 2            & 0              & 391                                                           & 55                                                           & $\checkmark$ \\
				CNN-ResNet         & 6         & ReLU       & 2            & 0              & 215                                                           & 41                                                           & $\checkmark$ \\
				CNN-DenseNet       & 6         & ReLU       & 2            & 1              & 163                                                           & 39                                                           & $\checkmark$ \\
				CNN-DenseNet-L1    & 6         & ReLU       & 2            & 1              & 1000                                                          & 39                                                           & $\checkmark$ \\
				CNN-DenseNet-L2    & 6         & ReLU       & 2            & 1              & 155                                                           & 39                                                           & $\checkmark$ \\
				CNN-ResNet-AvgPool & 6         & ReLU       & 2            & 0              & 109                                                           & 29                                                           & $\checkmark$ \\ \hline
			\end{tabular}%
		}
	\end{table}
	
	\section{Omitted Main Results: Neural Topology}
	\label{nta1}
	\subsection{Implementation Details}
	In neural topology, we analyze the geometry of neurons present in the hidden and the top layer.  Here, three different architectures with $2^{13}$, $2^{14}$ and $2^{15}$ hidden units are used to ensure sufficient expressive power. The core of our visualization is neural interaction which is modelled by Affinity Propagation (AP)~\cite{frey2006mixture,frey2007clustering}. Since each model has large number of neurons in the hidden layer, we restrict our topological analysis to a fixed subset of 2048 neurons. Due to extreme time and space complexity in AP, we first reduce the dimension of neurons in the hidden layer from $\mathbb{R}^{d_x}$ (here, $d_x = 784$) to $\mathbb{R}^{10}$ using PCA and thereafter, to $\mathbb{R}^{2}$ using t-SNE~\cite{maaten2008visualizing}. In case of top layer, we directly apply t-SNE to map neurons in $\mathbb{R}^{d_y}$ to $\mathbb{R}^{2}$ (here, $d_y = 10$). Note that the absolute units of x and y axes are not important in these neural topology diagrams. 
	
	\subsection{NTA on MNIST}
\begin{figure}[t]
		\centering
		\includegraphics[width=\columnwidth]{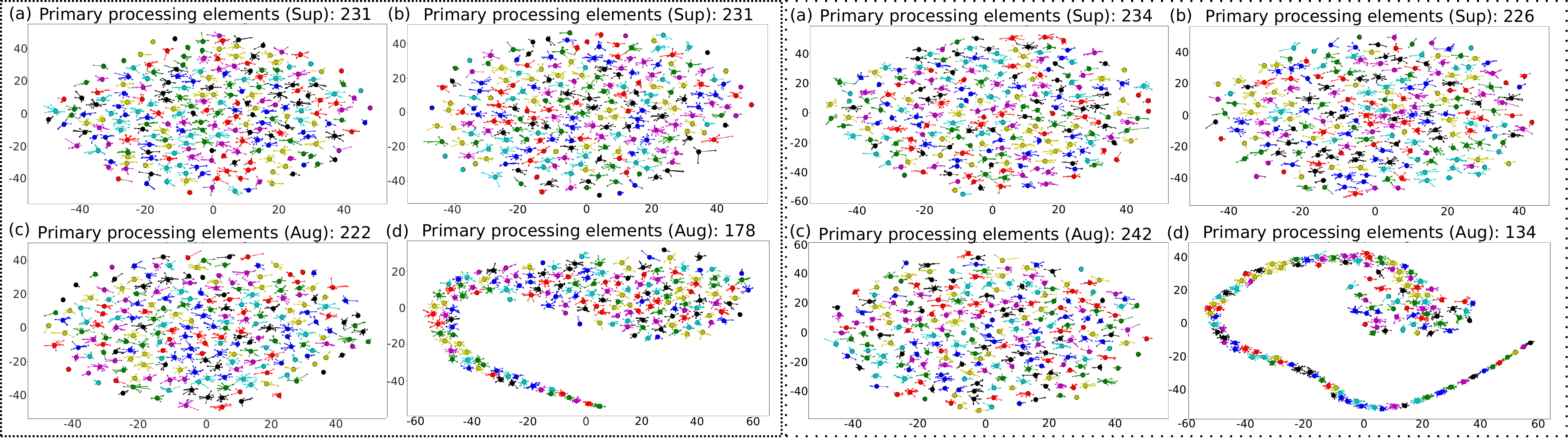}
		\caption{NTA in \textit{hidden layer} (left) and \textit{top layer} (right). (a) Initial and (b) final topology in supervised learning. (c) Initial and (d) final topology in adversarial learning.}
		\label{mnist_nta_hl_tl}
	\end{figure}

	In the experiments with $2^{14}$ hidden units, we observe emergence of evolutionary patterns in adversarial framework. As shown in Figure~\ref{mnist_nta_hl_tl}, even though both systems are initialized with similar topology in weight space, the final topology in regularized adversarial learning changes drastically. It is quite apparent from Figure~\ref{mnist_nta_hl_tl}(d), both in the hidden and the top layers, that adversarially learned weights lie on a different geometrical surface compared to sole supervision. Particularly intriguing is the self-organization tendency of these artificial neurons in a topological sense~\cite{kohonen1990self}. We observe this sparse self-organization behavior on a wide variety of architectures. In all these configurations, adversarial learning tries to exploit sparsity in data to reorganize neurons in neural embedded vector space.  
	
\begin{figure}[!t]
		\centering
		\includegraphics[width=\columnwidth]{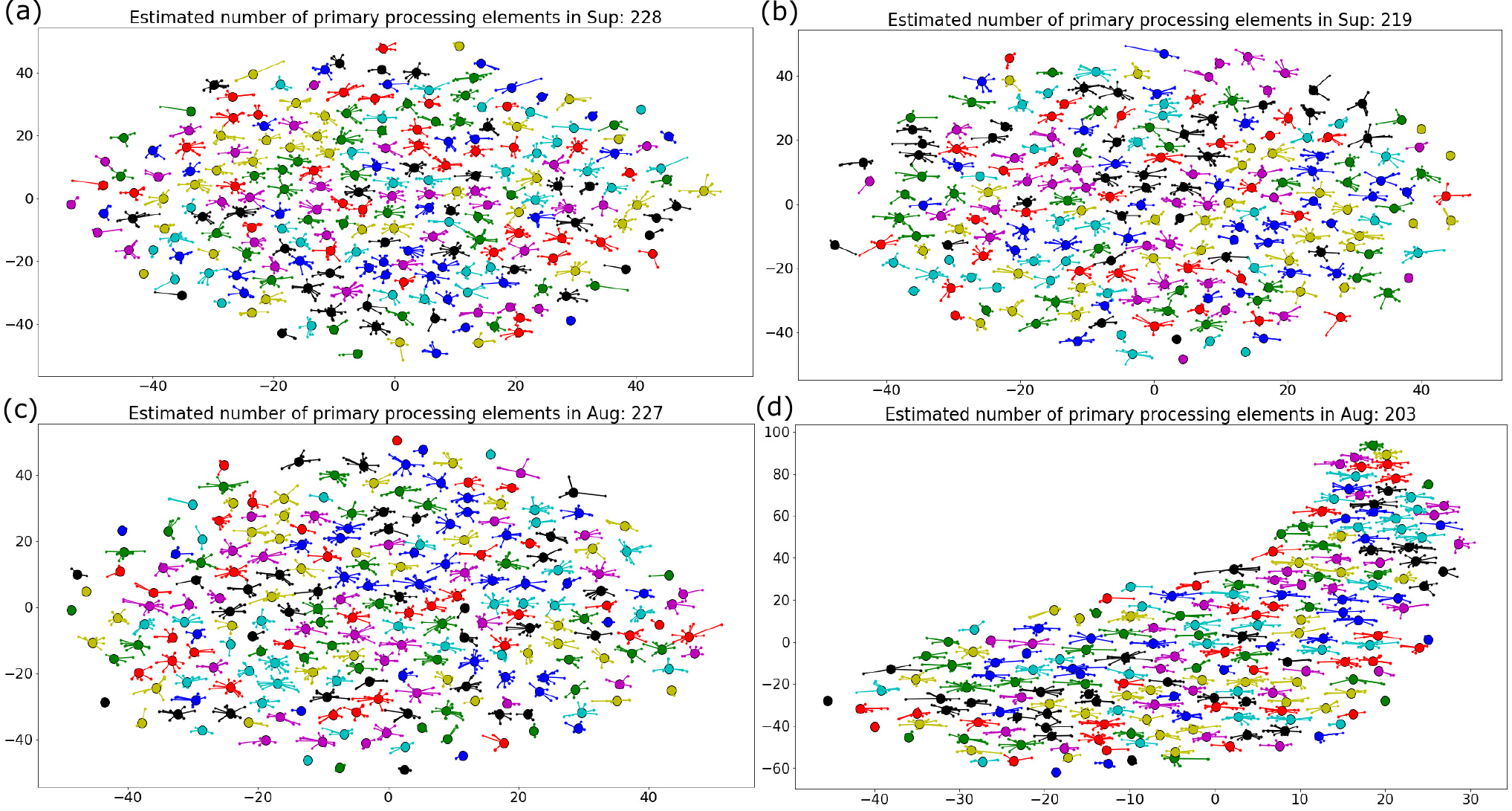}
		\caption{NTA in the \textit{hidden layer} with $2^{15}$ hidden units. (a) Initial and (b) final topology in supervised learning. (c) Initial and (d) final topology in adversarial learning.}
		\label{mnist_nta_hl_h15}
	\end{figure}

\begin{figure}[!t]
		\centering
		\includegraphics[width=\columnwidth]{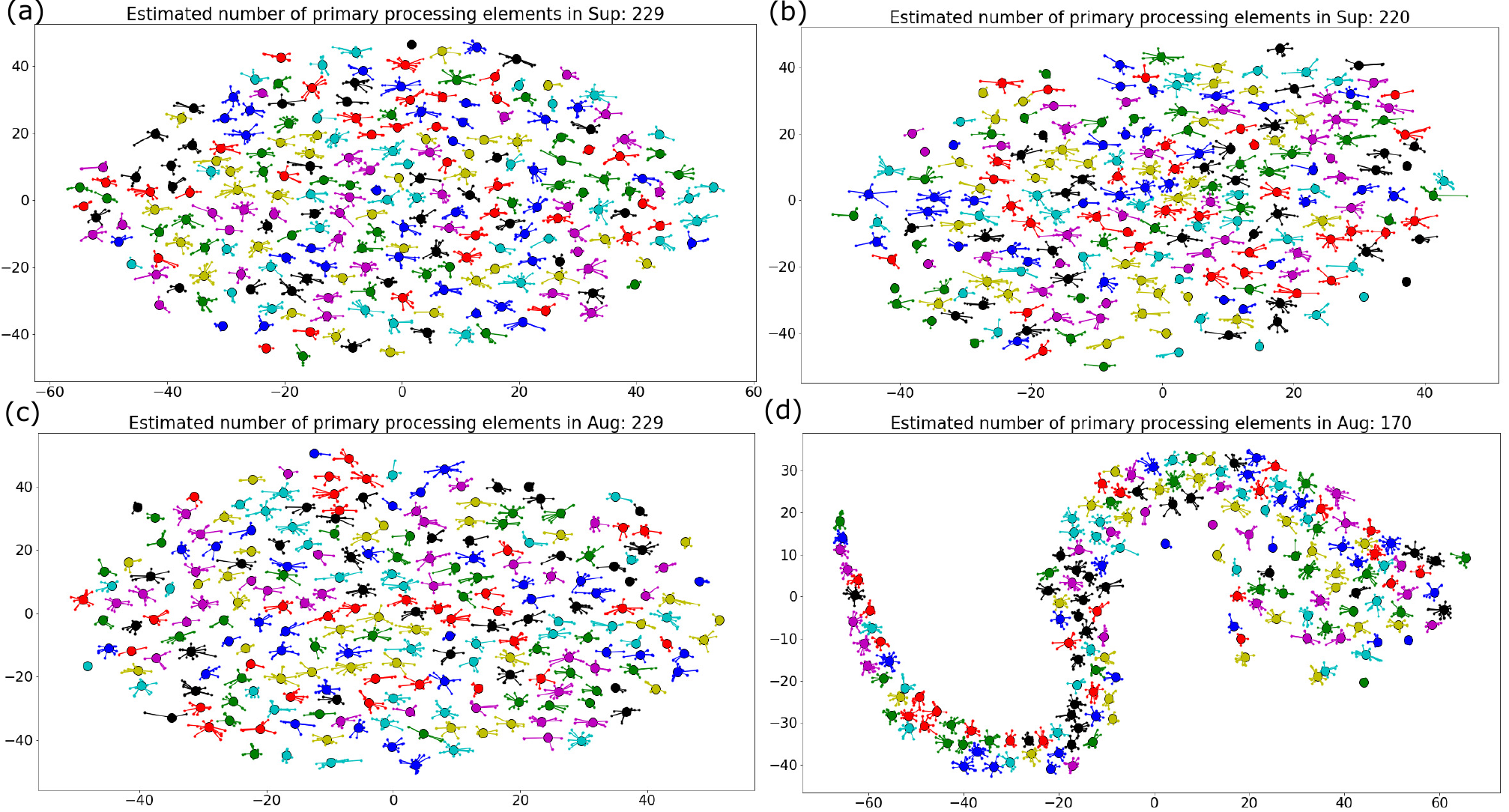}
		\caption{NTA in the \textit{top layer} with $2^{15}$ hidden units. (a) Initial and (b) final topology in supervised learning. (c) Initial and (d) final topology in adversarial learning.}
		\label{mnist_nta_tl_h15}
	\end{figure}

\begin{figure}[!t]
		\centering
		\includegraphics[width=\columnwidth]{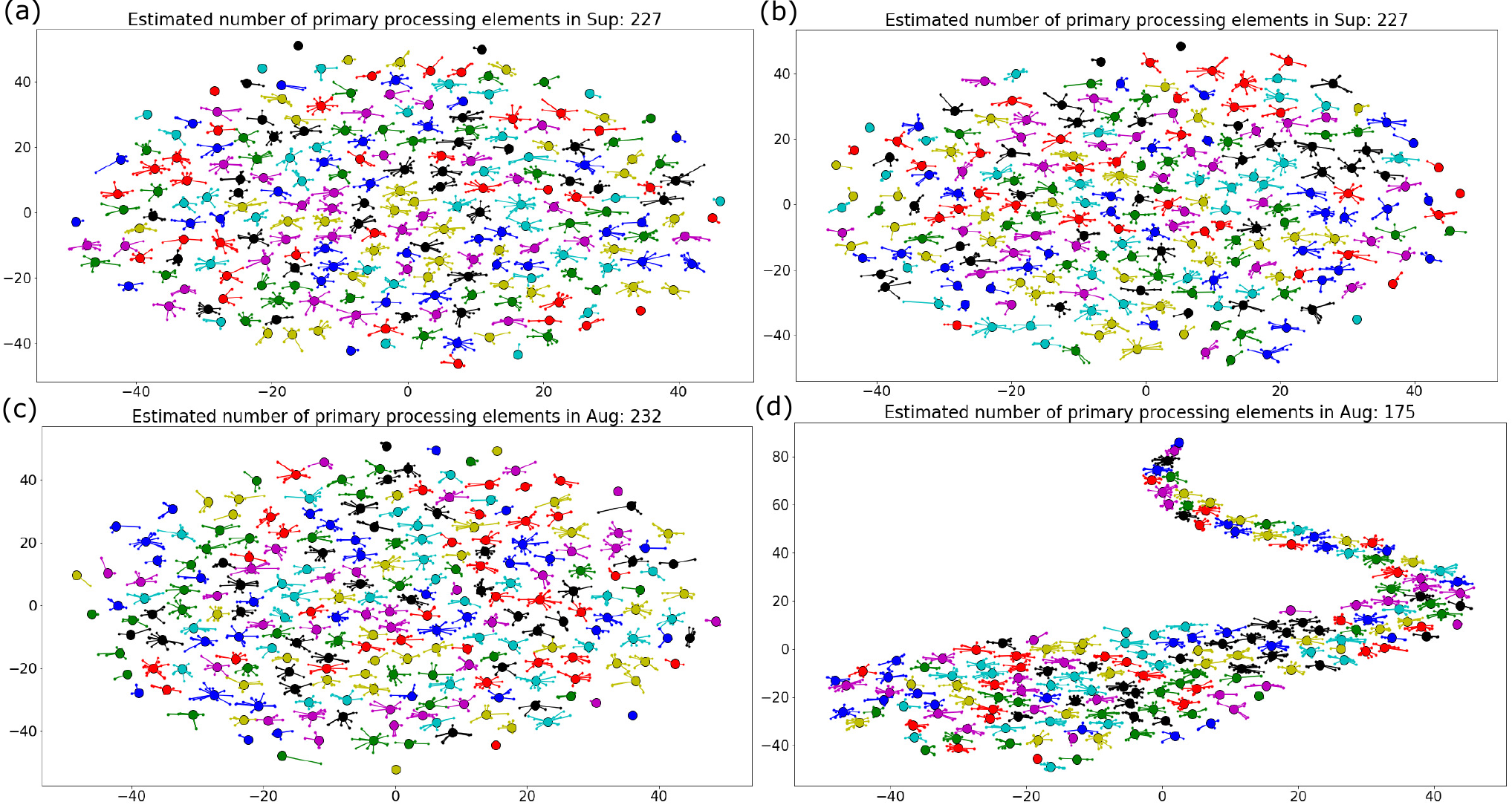}
		\caption{NTA in the \textit{hidden layer} with $2^{13}$ hidden units. (a) Initial and (b) final topology in supervised learning. (c) Initial and (d) final topology in adversarial learning.}
		\label{mnist_nta_hl_h13}
	\end{figure}

\begin{figure}[!t]
		\centering
		\includegraphics[width=\columnwidth]{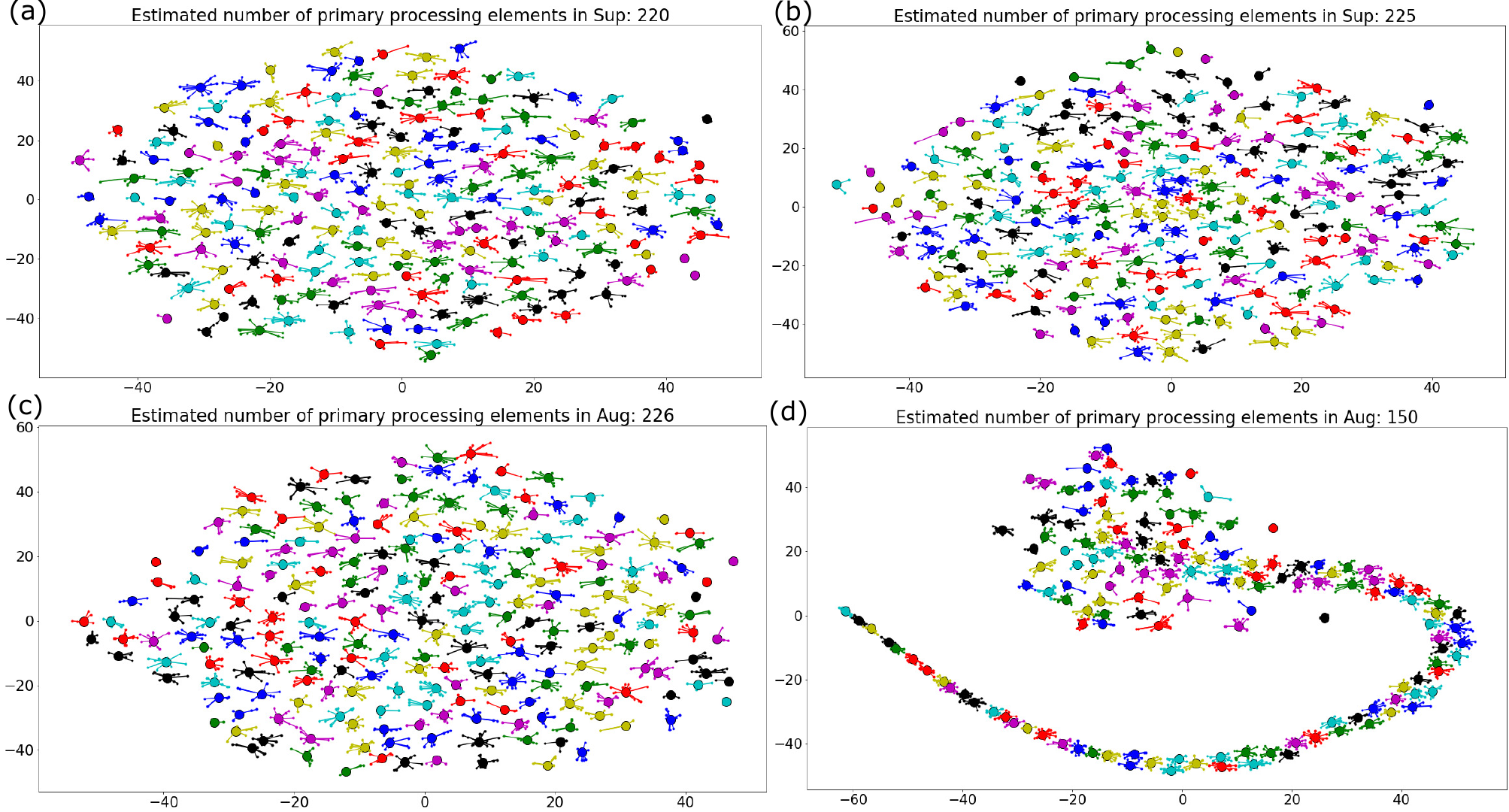}
		\caption{NTA in the \textit{top layer} with $2^{13}$ hidden units. (a) Initial and (b) final topology in supervised learning. (c) Initial and (d) final topology in adversarial learning.}
		\label{mnist_nta_tl_h13}
	\end{figure}

	In Figure~\ref{mnist_nta_hl_h15}~and~\ref{mnist_nta_tl_h15}, we also observe this drastic change in neural topology from initiation. The arguments are still supported in another architecture with $2^{15}$ hidden units on MNIST (Figure~\ref{mnist_nta_hl_h15}~and~\ref{mnist_nta_tl_h15}). Adversarial regularization exploits sparsity in data distribution and neural embedded topological vector space, provided it exists. This suggests participation of a smaller subset of neurons in achieving desired task. Vanishing gradient phenomenon, which we observed in both the layers, adds on to the explanation of lacking evolutionary patterns in sole supervision. Figure~\ref{mnist_nta_hl_h13}~and~\ref{mnist_nta_tl_h13} illustrate similar observations with $2^{13}$ hidden units on MNIST.

\begin{figure}[!t]
		\centering
		\includegraphics[width=\columnwidth]{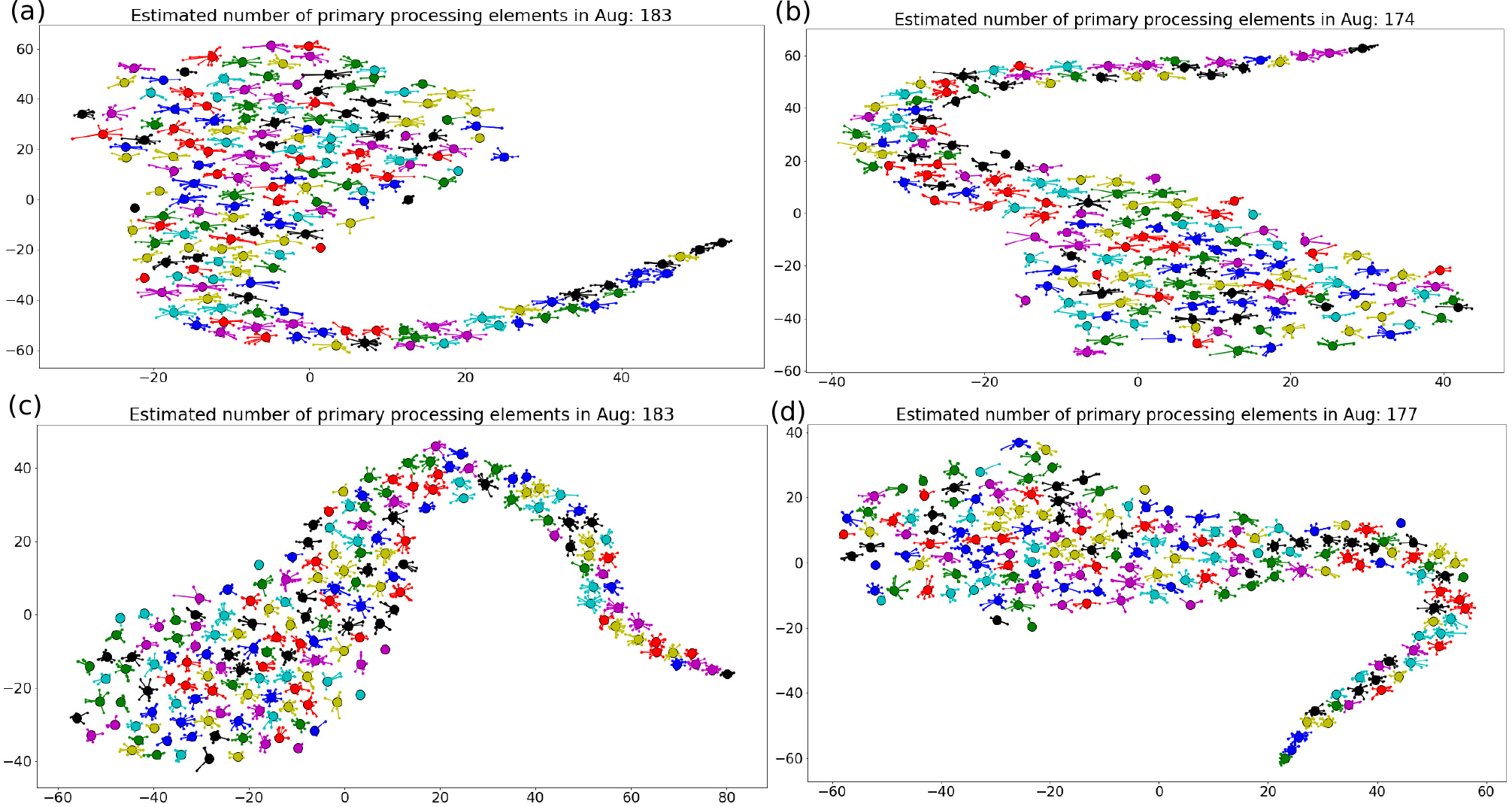}
		\caption{NTA in the \textit{hidden layer} with $2^{13}$ hidden units. (a) First subset (0-2048) (b) Second subset (2048-4096) (c) Third subset (4096-6144) (d) Fourth subset (6144-8192) final topology in adversarial learning.}
		\label{mnist_nta_hl_h13_sets}
	\end{figure}

\begin{figure}[!t]
		\centering
		\includegraphics[width=\columnwidth]{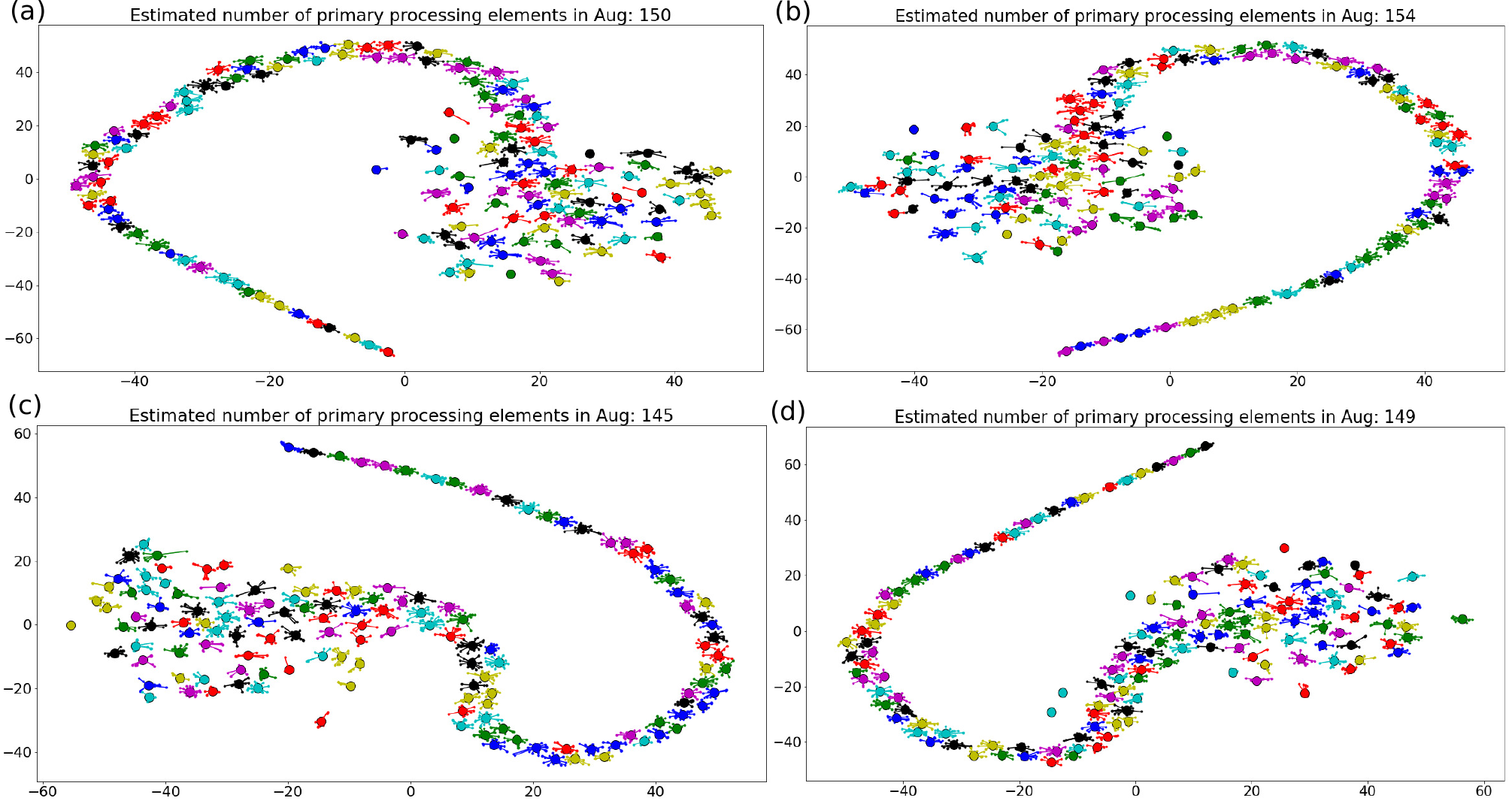}
		\caption{NTA in the \textit{top layer} with $2^{13}$ hidden units. (a) First subset (0-2048) (b) Second subset (2048-4096) (c) Third subset (4096-6144) (d) Fourth subset (6144-8192) final topology in adversarial learning.}
		\label{mnist_nta_tl_h13_sets}
	\end{figure}

\begin{figure}[!t]
		\centering
		\includegraphics[width=\columnwidth]{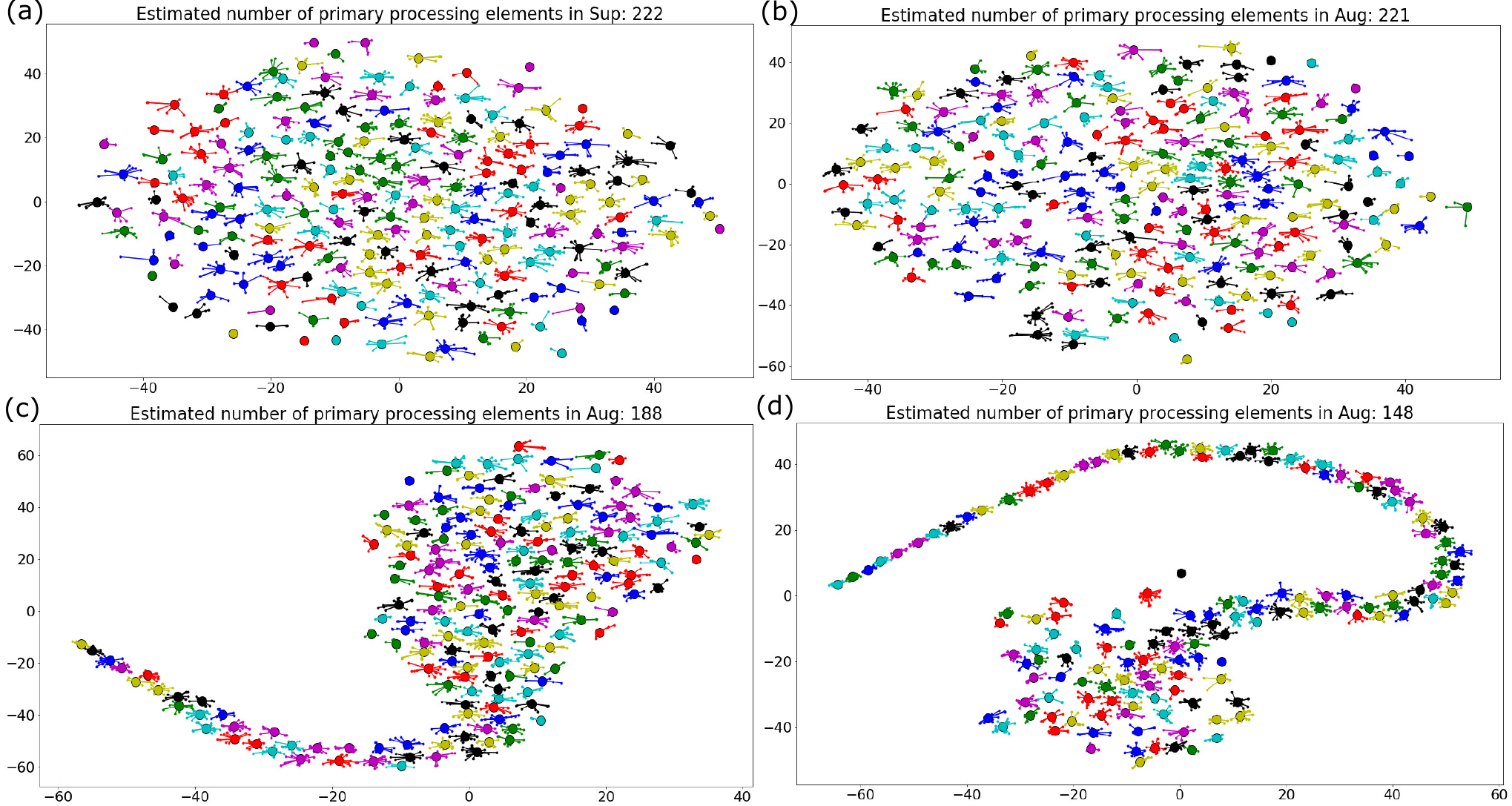}
		\caption{NTA of a random subset of 2048 neurons with $2^{13}$ hidden units.(a) Hidden and (b) top layer topology in supervised learning. (c) Hidden and (d) top layer topology in adversarial learning.}
		\label{mnist_fin_h13_rn}
	\end{figure}

\begin{figure}[!t]
		\centering
		\includegraphics[width=\columnwidth]{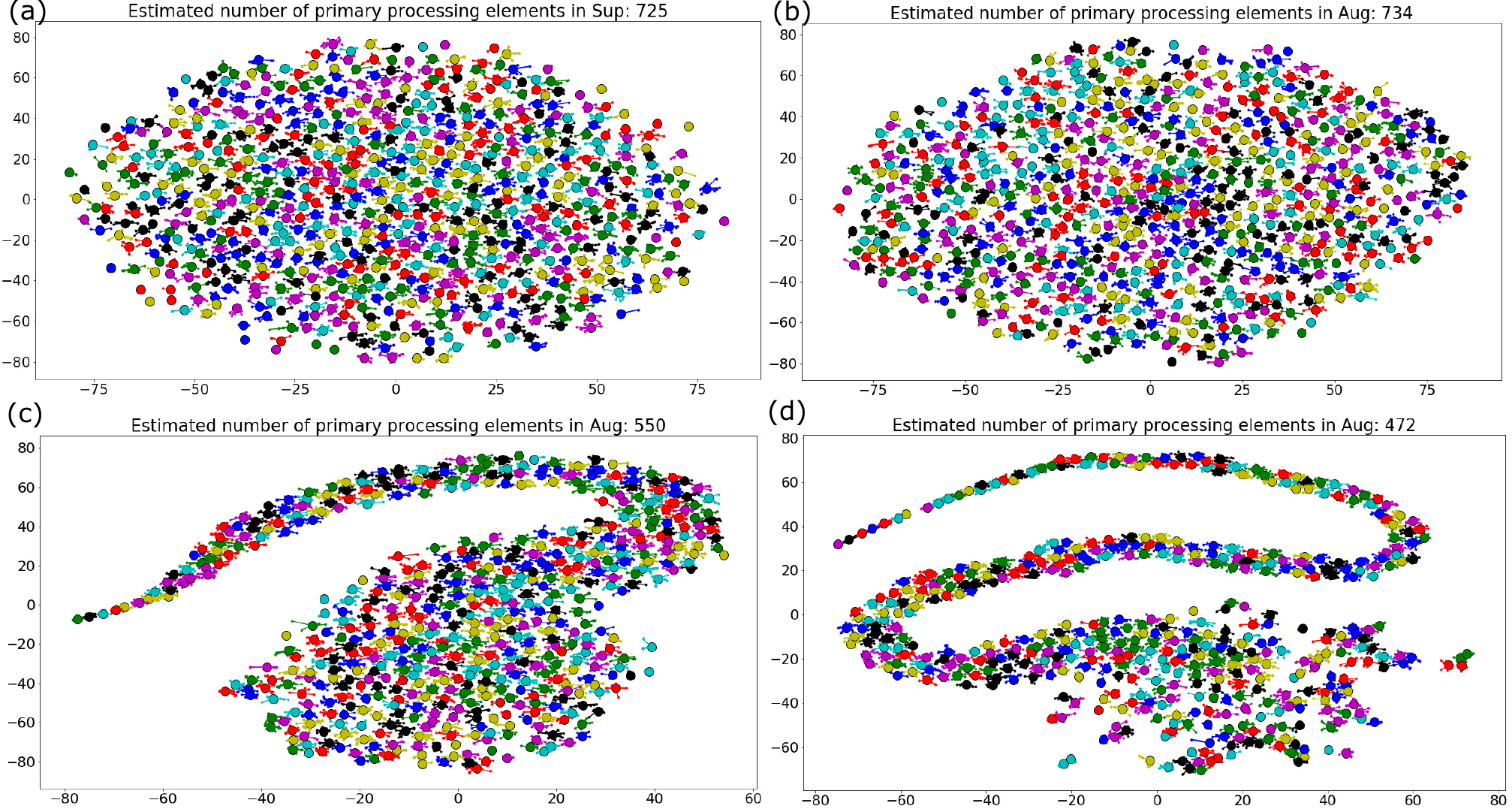}
		\caption{NTA of all $2^{13}$ hidden units.(a) Hidden and (b) top layer final topology in supervised learning. (c) Hidden and (d) top layer final topology in adversarial learning.}
		\label{mnist_fin_h13_all}
	\end{figure}

	Further, we study the neural topology of other fixed subsets of neurons in a network with $2^{13}$ hidden units as shown in Figure~\ref{mnist_nta_hl_h13_sets}~and~\ref{mnist_nta_tl_h13_sets}. In this analysis, we focus on 4 subsets of 2048 neurons each sequentially. Since we repeatedly observe new patterns even with random seeds, it ensures that the geometry of neural embedded vector space has indeed changed drastically. Also, we analyze the topology of a randomly selected subset of 2048 neurons with $2^{13}$ hidden units. As shown in Figure~\ref{mnist_fin_h13_rn}, the final topology in adversarial learning does lie on a different manifold as compared to sole supervision. In addition, Figure~\ref{mnist_fin_h13_all} shows emergence of \textit{global pattern} in adversarial learning due to more local interaction~\cite{turing1952chemical}.

	\subsubsection{Perturbation Sensitivity}
\begin{figure}[!t]
		\centering
		\includegraphics[width=\columnwidth]{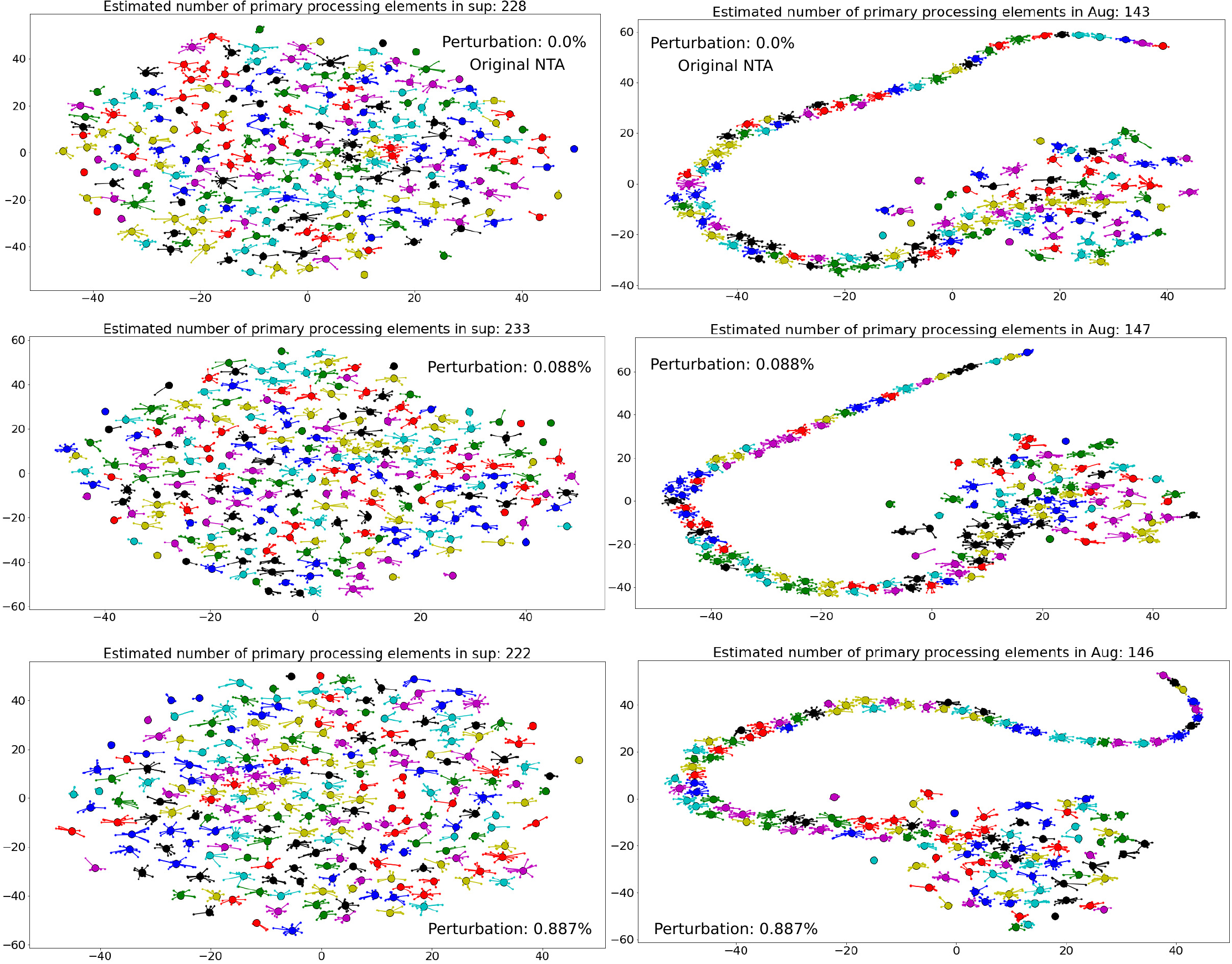}
		\caption{NTA in the \textit{top layer} with $2^{13}$ hidden units. Comparison of sensitivity to low level Gaussian perturbation. Final topology in supervised learning (left) and adversarial learning (right).}
		\label{perturb}
	\end{figure}

\begin{figure}[!t]
		\centering
		\includegraphics[width=\columnwidth]{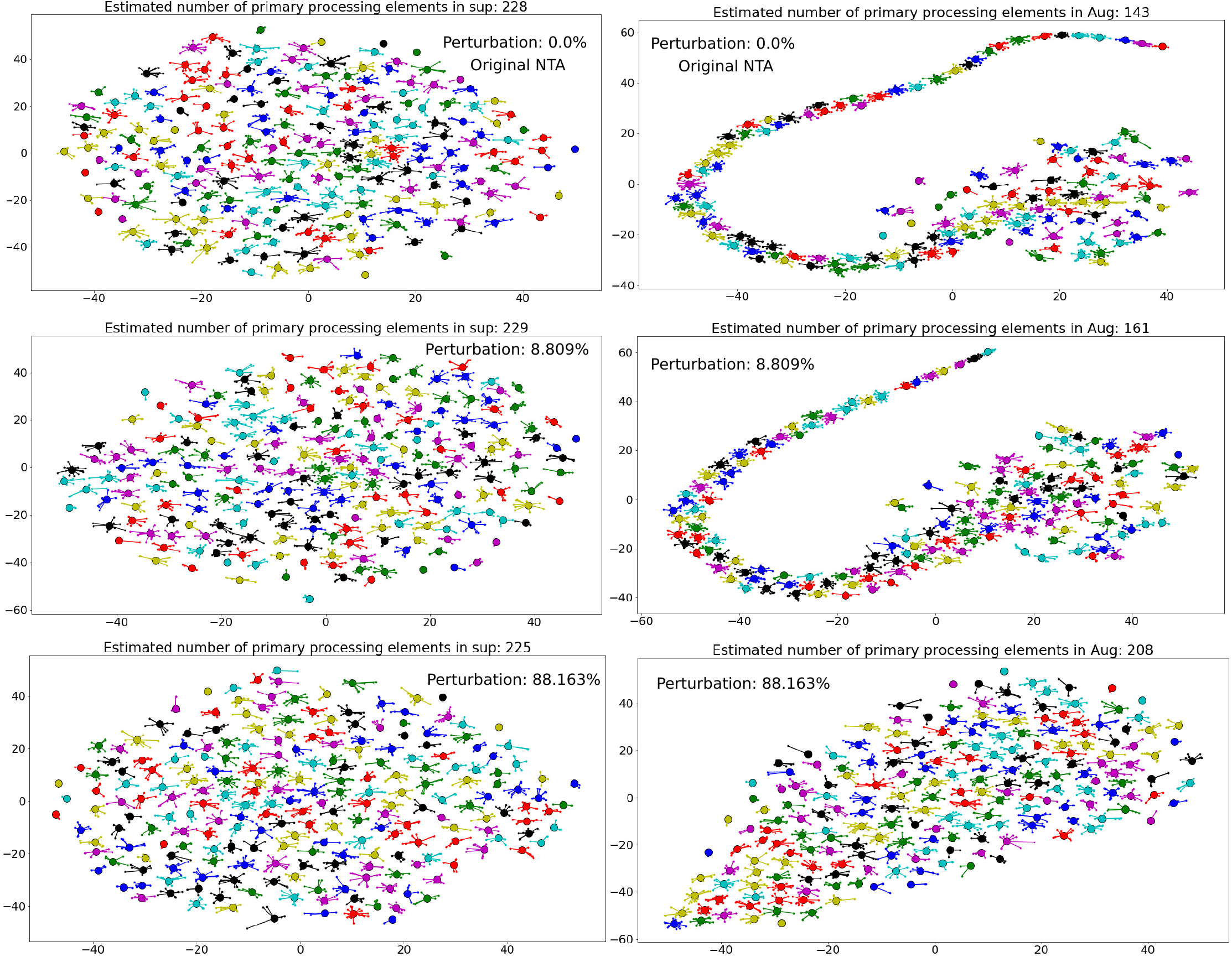}
		\caption{NTA in the \textit{top layer} with $2^{13}$ hidden units. Comparison of sensitivity to moderate and extreme level Gaussian perturbation. Final topology in supervised learning (left) and adversarial learning (right).}
		\label{perturb1}
	\end{figure}

	In Figure~\ref{perturb}~and~\ref{perturb1}, we investigate the sensitivity of the topological diagrams to local perturbation. The perturbation model considered here follows Gaussian distribution with mean and standard deviation same as that of the fully trained weights. Here, the percentage perturbation corresponds to the fraction of the total energy in the weight vectors. For conciseness, we study sensitivity in the top layer on MNIST with $2^{13}$ hidden nodes. As shown in Figure~\ref{perturb} and~\ref{perturb1}, the final topology retains sparse representation with low and moderate level Gaussian perturbation. However, we observe slight reduction of sparsity with extreme perturbation as shown in Figure~\ref{perturb1}. These experimental results indicate that the sparse nature of neural anatomy in augmented objective is not due to minor deviations from the neural anatomy of sole supervision. Thus, there is a significant difference between the final topology of adversarial regularization and sole supervision in the neural embedded vector space.

	\subsection{NTA on Over-Parameterization}
	It is well known that highly over-parameterized deep neural networks sprinkle the corresponding parametric space with lots of good solutions. However, it is not fully understood how to reduce this dependency on over-parameterization while still achieve required performance. In this paper, we illustrate this phenomenon using topological diagrams of fully trained networks. The fact that all neurons in the weight space do not contribute equally to the main task highlights the existence of redundant neurons in over-parameterized networks --- though in a positive sense.
	
\begin{figure}[!t]
		\centering
		\includegraphics[width=\columnwidth]{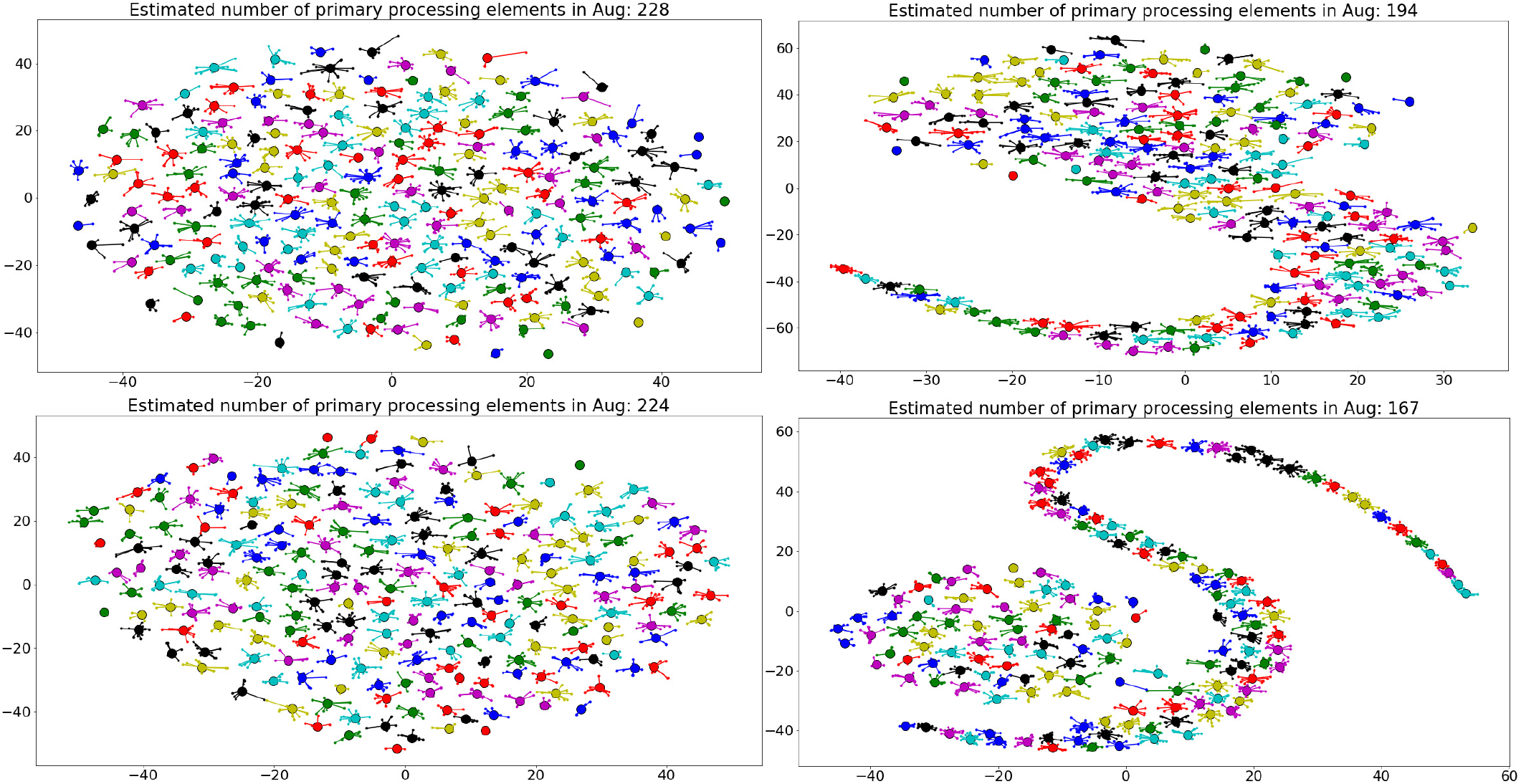}
		\caption{NTA in \textit{adversarial learning} with $2^{13}$ hidden units. Initial and final topology in \textit{hidden layer} (first row) and \textit{top layer} (second row). }
		\label{rl}
	\end{figure}

	In Figure~\ref{rl}, we study the neural topology of a network which is trained on randomly labelled pairs of MNIST dataset. With $2^{13}$ nodes in hidden layer, the augmented objective converges to 0.004 MSE after 1000 epochs. It is interesting to observe these patterns even when trained on a randomly labelled dataset. This purportedly implies that adversarial training is the predominant source that constitutes the basis of such pattern formation.

	\subsection{NTA on FashionMNIST}
\begin{figure}[!t]
		\centering
		\includegraphics[width=\columnwidth]{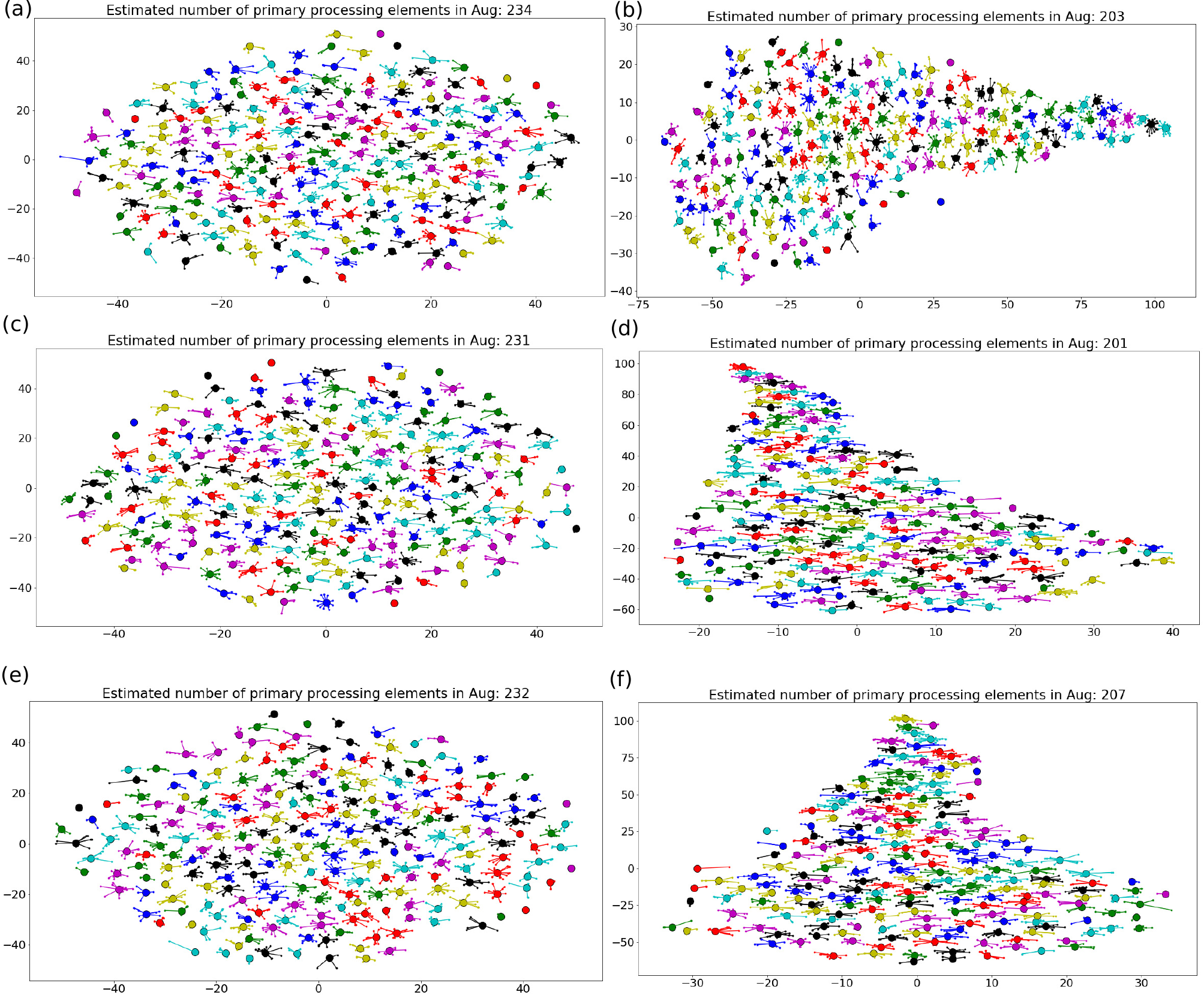}
		\caption{NTA in the \textit{hidden layer} with $2^{13}$ hidden units on FashionMNIST. (a) First subset (0-2048) (b) Second subset (2048-4096) (c) Third subset (4096-6144) initial (left) and final (right) topology in adversarial learning.}
		\label{fmnist_nta_hl_h13_sets}
	\end{figure}

\begin{figure}[!t]
		\centering
		\includegraphics[width=\columnwidth]{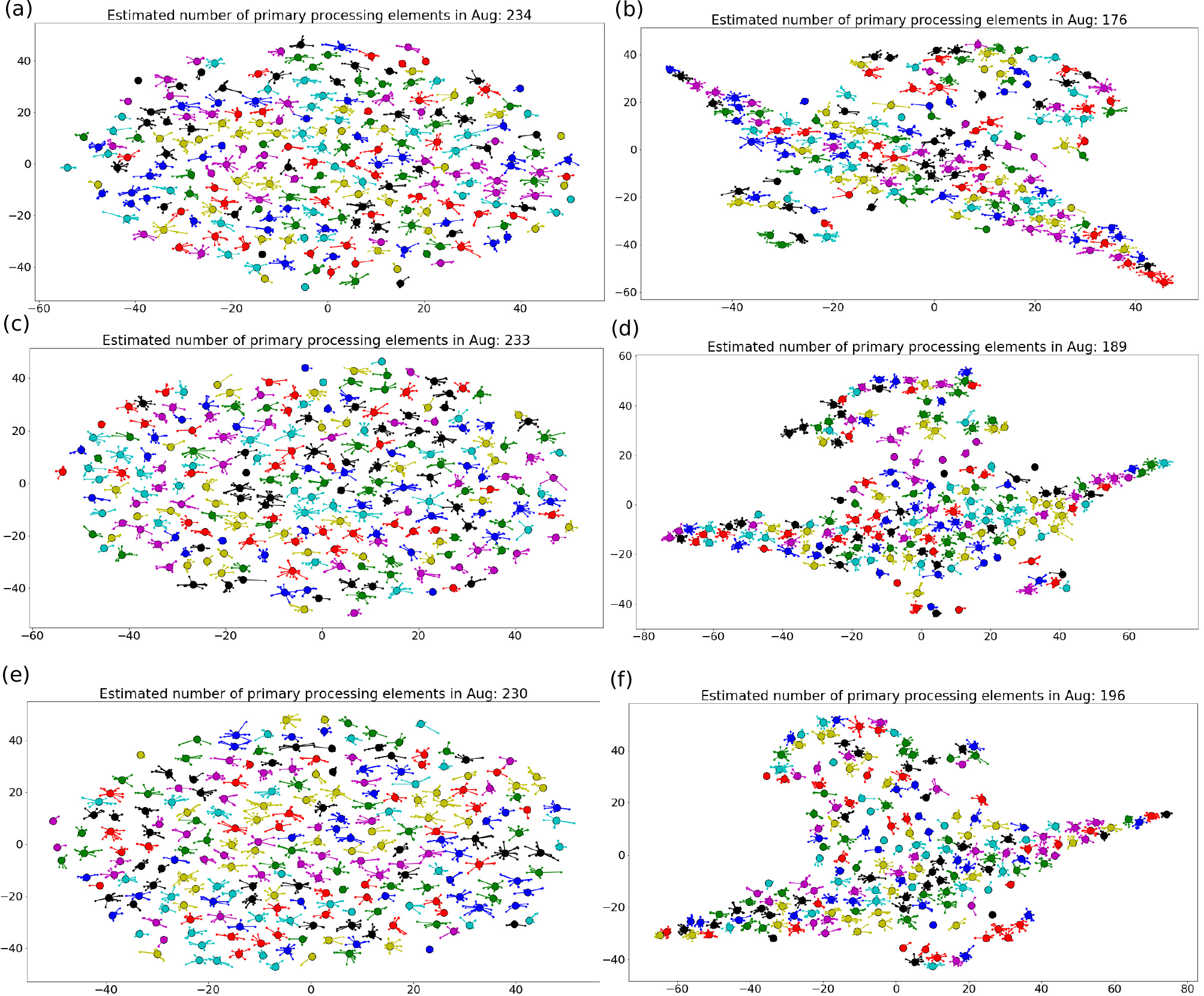}
		\caption{NTA in the \textit{top layer} with $2^{13}$ hidden units on FashionMNIST. (a) First subset (0-2048) (b) Second subset (2048-4096) (c) Third subset (4096-6144) initial (left) and final (right) topology in adversarial learning.}
		\label{fmnist_nta_tl_h13_sets}
	\end{figure}

	Additionally, the experiments on FashionMNIST demonstrate similar pattern formation on three different subsets as shown in Figure~\ref{fmnist_nta_hl_h13_sets}~and~\ref{fmnist_nta_tl_h13_sets}.

	\subsection{Neural Anatomy}
	Since we believe all neurons lie on a common manifold due to single channel representation of input data, it makes more sense to study their topology on single channel datasets, such as MNIST and FashionMNIST. However, there are several potential improvements and extensions to the present account of neural topology analysis. A particularly interesting research direction would be to design an experiment for multi-channel dataset, such as SVHN and CIFAR10. We believe that studying channel specific topology might give insights to design better architectures. Also, it is worth unveiling whether there exist such patterns in convolutional neural networks. 
	
	An interesting observation in most of these diagrams is the emergence of animal shaped patterns with central and assistant nervous systems. \citeauthor{turing1952chemical}'s theory predicted that the emergence of patterns on the skin of an animal is due to chemical substances, called morphogens reacting together and diffusing through tissues~\cite{turing1952chemical}. In the context of morhphogenesis, while one reaction favors the growth of patterns, another tries to prohibit it. In the nascent state of understanding, this forms the chemical basis of morphogenesis. To our surprise, the adversarial game between generator and discriminator also forms a similar basis for the evolutionary pattern formation in neural topology, suggesting further research in this direction might prove beneficial.

	The neural topology in adversarial regularization has essentially two components: a central nervous system/dense branch and an assistant nervous system/narrow branch. The resemblance of dense branch with neural topology diagram in sole supervision suggests that adversarial learning somehow exploits sparsity in over-parameterized neural networks. Furthermore, it provides accelerated gradients in the optimization process. As it turns out, adversarial learning depends upon a very few primary processing elements to efficiently perform the same task. It is however unclear at the moment the exact role of each of these individual branches. It makes one wonder whether local neural interaction, which is believed to be the primary cause of such evolutionary patterns, can help in reasoning, interpretability and designing efficient architectures upon further investigation.
	
	To our knowledge, one can not at present hope to make progress in understanding the electrical, chemical and mechanical properties of neurons in the fabric of space and time that influence the emergence of evolutionary patterns. It is hoped, however, that the simplified architectures retained for discussion are those of greatest importance at this juncture. Thus, the present account of the problem is vastly a simplification and an idealization of actual neural anatomy. It is intended to bridge the gap between chemical basis of morphogenesis and an equivalent mathematical basis of neural topology.
	
	\section{Technical Proofs}
	\subsection{Proof of Theorem 2}
	\label{pr_th2}
	We parameterize the path between $\theta_k$ and $\theta_{k+1}$ as following:
	\begin{equation}
	\gamma(t) = t\theta_{k+1} + (1-t)\theta_k \forall t\in [0,1].
	\end{equation}
	By fixed step gradient descent, the iterate $\theta_{k+1} = \theta_k - h_k \nabla l(\theta_k)$. Using Taylor's expansion,
	\begin{equation}
	\begin{split}
	l\left ( \theta_{k+1} \right ) &= l\left ( \theta_{k} \right ) + \nabla l\left ( \theta_{k} \right ) \left ( \theta_{k+1} - \theta_k \right ) + \frac{1}{2} \left ( \theta_{k+1} - \theta_k \right )^T \nabla^2 l\left ( \theta_{k} \right )\left ( \theta_{k+1} - \theta_k \right )\\
	& = l\left ( \theta_{k} \right ) - h_k\left \| \nabla l\left ( \theta_{k} \right )  \right \|^2 + \frac{1}{2} \left ( \theta_{k+1} - \theta_k \right )^T \nabla^2 l\left ( \theta_{k} \right )\left ( \theta_{k+1} - \theta_k \right ),~\left ( \because \theta_{k+1} - \theta_k = -h_k \nabla l(\theta_k) \right ).
	\end{split}
	\end{equation}
	Using Cauchy-Schwarz inequality and integrating over parameterized curve $\gamma(t)$, 
	\begin{equation}
	\begin{split}
	l\left ( \theta_{k+1} \right ) & \leq l\left ( \theta_{k} \right ) - h_k\left \| \nabla l\left ( \theta_{k} \right )  \right \|^2 + \frac{1}{2} \left \| \left ( \theta_{k+1} - \theta_k \right ) \right \| \left \| \nabla^2 l\left ( \theta_{k} \right )\left ( \theta_{k+1} - \theta_k \right ) \right \| \\
	& \leq l\left ( \theta_{k} \right ) - h_k\left \| \nabla l\left ( \theta_{k} \right )  \right \|^2 + \frac{1}{2} \left \| \left ( \theta_{k+1} - \theta_k \right ) \right \|^2 \int_{0}^{1} \left \| \nabla^2 l\left ( \gamma(t) \right ) \right \| dt.
	\end{split}
	\end{equation}
	
	We know by \textbf{Assumption 5}
	\begin{equation}
	\left \| \nabla^2l\left ( \theta \right ) \right \| \leq L_0 + L_1 \left \| \nabla l\left ( \theta \right ) \right \|.
	\end{equation}
	Then using descent rule and arguments of \textbf{Theorem 1}, we obtain the following inequality:
	\begin{equation}
	\begin{split}
	l\left ( \theta_{k+1} \right ) & \leq l\left ( \theta_{k} \right ) - h_k\left \| \nabla l\left ( \theta_{k} \right )  \right \|^2 + \frac{h_k^2 \left \|\nabla l\left ( \theta_{k} \right )  \right \|^2 }{2} \int_{0}^{1} \left ( L_0 + L_1\left \| \nabla l\left ( \gamma(t) \right ) \right \| \right ) dt \\
	& \leq l\left ( \theta_{k} \right ) - h_k\left \| \nabla l\left ( \theta_{k} \right )  \right \|^2 + \frac{h_k^2 \left \|\nabla l\left ( \theta_{k} \right )  \right \|^2 }{2} \int_{0}^{1} \left ( L_0 + L_1 L^2 \beta \epsilon \right ) dt\\
	& \leq l\left ( \theta_{k} \right ) - h_k\left \| \nabla l\left ( \theta_{k} \right )  \right \|^2 + \frac{h_k^2 \left \|\nabla l\left ( \theta_{k} \right )  \right \|^2 \left ( L_0 + L_1L^2 \beta \epsilon\right ) }{2} .
	\end{split}
	\end{equation}
	Let us choose $h_k = \frac{1}{L_0 + L_1 L^2 \beta \epsilon }$. Now,
	\begin{equation}
	\begin{split}
	l\left ( \theta_{k+1} \right ) & \leq l\left ( \theta_{k} \right ) -  \frac{h_k \left \|\nabla l\left ( \theta_{k} \right )  \right \|^2 }{2} \\
	& \leq l\left ( \theta_{k} \right ) -  \frac{\left \|\nabla l\left ( \theta_{k} \right )  \right \|^2 }{2\left ( L_0 + L_1 \lambda M \right )} .
	\end{split}
	\end{equation}
	Assume that it takes $T$ iterations to reach $\epsilon$-stationary point, i.e., $\epsilon \leq \left \| \nabla l\left ( \theta_{k} \right ) \right \|$ for $k \leq T$. By a telescopic sum over $k$,
	\begin{equation}
	\begin{split}
	\sum_{k=0}^{T-1}l\left ( \theta_{k+1} \right ) - l\left ( \theta_k \right ) & \leq \frac{-T\epsilon^2}{2\left ( L_0+L_1\lambda M \right )}\\
	\implies T &\leq \frac{2 \left ( l(\theta_0) - l^* \right ) \left ( L_0+L_1 L^2 \beta \epsilon \right )}{\epsilon^2}.
	\end{split}
	\end{equation}
	Therefore, we get
	\begin{equation}
	\sup_{\theta_0 \in \left \{ \mathbb{R}^{h\times d_x}, \mathbb{R}^{d_y\times h} \right \},l\in \mathscr{L}}\mathcal{T}_\epsilon\left (A_h \left [ l ,\theta_0 \right ],l  \right ) = \mathcal{O}\left ( \frac{\left ( l(\theta_0)- l^* \right ) \left ( L_0 + L_1 L^2 \beta \epsilon \right )}{\epsilon^2} \right )
	\end{equation}
	which finishes the proof.\hfill$\square$
	
	\subsection{Proof of Corollary 1}
	\label{pr_co1}
	Using the arguments made in the proof of \textbf{Theorem 2} and first-order Taylor's expansion, we get
	\begin{equation}
	\begin{split}
	l\left ( \theta_{k+1} \right ) & = l\left ( \theta_k \right ) - h_k\left \| \nabla l\left ( \theta_k \right ) \right \|^2\\
	&\leq l\left ( \theta_k \right ) - h_k\epsilon^2.
	\end{split}
	\end{equation}
	By telescopic sum, 
	\begin{equation}
	\begin{split}
	\sum_{k=0}^{T-1}  l\left ( \theta_{k+1}\right ) - l\left ( \theta_k \right ) &\leq -T h_k \epsilon^2\\
	\implies T &\leq \frac{\left ( l\left ( \theta_{0}\right ) - l^* \right )}{h_k \epsilon^2}.
	\end{split}
	\end{equation}
	So, 
	\begin{equation}
	\sup_{\theta_0 \in \left \{ \mathbb{R}^{h\times d_x}, \mathbb{R}^{d_y\times h} \right \},l\in \mathscr{L}}\mathcal{T}_\epsilon\left (A_h \left [ l ,\theta_0 \right ],l  \right ) = \mathcal{O}\left ( \frac{\left ( l\left ( \theta_{0}\right ) - l^* \right )}{h \epsilon^2} \right )
	\end{equation}
	which finishes the proof.\hfill$\square$

	\subsection{Proof of Theorem 3}
	\label{pr_th3}
	Recall that the target function $l(\theta)$ remains identical in both settings except for additional cost of discriminator over generator in augmented objective. In this setting, the parameters are updated as 
	\begin{equation}
	\theta_{k+1} = \theta_k - h_k \nabla \left ( l\left ( \theta_k \right ) - g \left (\psi; f\left (\theta_k; x \right ) \right ) \right ).
	\end{equation}
	Using Taylor's expansion, the triangle and Cauchy-Schwarz inequality as in Appendix~\ref{pr_th2}, we obtain
	\begin{equation}
	\begin{split}
	l\left ( \theta_{k+1} \right ) &\leq l\left ( \theta_{k} \right ) - h_k \left \| \nabla l\left ( \theta_{k} \right ) \right \|^2 - h_k\left \| \nabla l\left ( \theta_{k} \right ) \right \|\left \| \nabla g \left (\psi; f\left (\theta_k; x \right ) \right ) \right \|  + \frac{h_k^2\left \| \nabla \left ( l\left ( \theta_k \right ) - g \left (\psi; f\left (\theta_k; x \right ) \right ) \right ) \right \|^2}{2}\int_{0}^{1}\left \| \nabla^2l(\gamma(t)) \right \| dt.
	\end{split}
	\end{equation}
	By \textbf{Assumption 5} and \textbf{6},
	\begin{equation}
	\begin{split}
	l\left ( \theta_{k+1} \right ) \leq l\left ( \theta_{k} \right ) - h_k \left \| \nabla l\left ( \theta_{k} \right ) \right \|^2 - h_k\left \| \nabla l\left ( \theta_{k} \right ) \right \| \zeta  + \frac{h_k^2\left \| \nabla l\left ( \theta_k \right ) - \nabla g \left (\psi; f\left (\theta_k; x \right ) \right ) \right \|^2}{2}\int_{0}^{1}\left ( L_0 + L_1 \left \| \nabla l(\gamma(t)) \right \| \right ) dt.
	\end{split}
	\end{equation}
	Upon simplification using arguments of Appendix~\ref{pr_th2} and applying Minkowski's inequality,
	\begin{equation}
	\begin{split}
	l\left ( \theta_{k+1} \right ) \leq l\left ( \theta_{k} \right ) - h_k \left \| \nabla l\left ( \theta_{k} \right ) \right \|^2 - h_k\left \| \nabla l\left ( \theta_{k} \right ) \right \| \zeta  + \frac{h_k^2\left ( \left \| \nabla l\left ( \theta_k \right )\right \|^2 + \left \| \nabla g \left (\psi; f\left (\theta_k; x \right ) \right ) \right \|^2 \right )}{2}\left ( L_0 + L_1 \lambda M \right ).
	\end{split}
	\end{equation}
	Using $h_k = \frac{1}{L_0 + L_1 L^2 \beta \epsilon}$, we get
	\begin{equation}
	\begin{split}
	l\left ( \theta_{k+1} \right ) &\leq l\left ( \theta_{k} \right ) - \frac{h_k \left \| \nabla l\left ( \theta_{k} \right ) \right \|^2}{2} - h_k\left \| \nabla l\left ( \theta_{k} \right ) \right \| \zeta  + \frac{h_k\left \| \nabla g \left (\psi; f\left (\theta_k; x \right ) \right ) \right \|^2 }{2}\\
	& \leq l\left ( \theta_{k} \right ) - \frac{h_k \left \| \nabla l\left ( \theta_{k} \right ) \right \|^2}{2} - h_k\left \| \nabla l\left ( \theta_{k} \right ) \right \| \zeta  + \frac{h_k L^2 \delta^2 }{2},~(\text{from \textbf{Lemma 2}}).
	\end{split}
	\end{equation}
	Assuming $T$ iterations to reach $\epsilon$-stationary point, i.e., $\epsilon \leq \left \| \nabla l\left ( \theta_{k} \right ) \right \|$ for $k \leq T$. By a telescopic sum over $k$,
	\begin{equation}
	\begin{split}
	\sum_{k=0}^{T-1}l\left ( \theta_{k+1} \right ) - l\left ( \theta_k \right ) &\leq \frac{-T\left ( \epsilon^2+2\epsilon\zeta - L^2 \delta^2  \right )}{2\left ( L_0+L_1  L^2 \beta \epsilon \right )}\\
	\implies T & \leq \frac{2 \left ( l(\theta_0) - l^* \right ) \left ( L_0+L_1  L^2 \beta \epsilon \right )}{\epsilon^2+2\epsilon\zeta - L^2 \delta^2}.
	\end{split}
	\end{equation}
	Therefore, we obtain
	\begin{equation}
	\sup_{\theta_0 \in \left \{ \mathbb{R}^{h\times d_x}, \mathbb{R}^{d_y\times h} \right \},l\in \mathscr{L}}\mathcal{T}_\epsilon\left (A_h \left [ l ,\theta_0 \right ],l  \right ) = \mathcal{O}\left ( \frac{\left ( l(\theta_0)- l^* \right ) \left ( L_0+L_1\lambda M \right )}{\epsilon^2 + 2\epsilon \zeta - \delta^2 M^2} \right )
	\end{equation}
	which finishes the proof.\hfill$\square$
	
	\subsection{Proof of Corollary 2}
	\label{pr_co2}
	Using the arguments made in the proof of \textbf{Theorem 3} and first-order Taylor's approximation, we get
	\begin{equation}
	\begin{split}
	l\left ( \theta_{k+1} \right ) & = l\left ( \theta_k \right ) - h_k\left \| \nabla l\left ( \theta_k \right ) \right \|^2 - h_k\left \|\nabla l\left ( \theta_k \right )  \right \| \left \| \nabla  g \left (\psi; f\left (\theta_k; x \right ) \right ) \right \|\\
	&\leq l\left ( \theta_k \right ) - h_k\epsilon^2 - h_k \epsilon \zeta.
	\end{split}
	\end{equation}
	By telescopic sum, 
	\begin{equation}
	\begin{split}
	\sum_{k=0}^{T-1}  l\left ( \theta_{k+1}\right ) - l\left ( \theta_k \right ) &\leq -T h_k \epsilon^2  - T h_k \epsilon \zeta\\
	\implies T &\leq \frac{\left ( l\left ( \theta_{0}\right ) - l^* \right )}{h_k \epsilon^2 + h_k \epsilon \zeta}.
	\end{split}
	\end{equation}
	Therefore, 
	\begin{equation}
	\sup_{\theta_0 \in \left \{ \mathbb{R}^{h\times d_x}, \mathbb{R}^{d_y\times h} \right \},l\in \mathscr{L}}\mathcal{T}_\epsilon\left (A_h \left [ l ,\theta_0 \right ],l  \right ) = \mathcal{O}\left ( \frac{\left ( l\left ( \theta_{0}\right ) - l^* \right )}{h \epsilon^2 + h \epsilon \zeta} \right )
	\end{equation}
	which finishes the proof.\hfill$\square$

	\subsection{Proof of Theorem 4}
	\label{pr_th4}
	In sole supervision, the parameters are updated by $\frac{d\theta(t)}{dt} = - \nabla l(\theta(t))$. We define distance to optimal solution as $r^2(t) = \frac{1}{2} \left \| \theta(t)-\theta^* \right \|^2$. Now differentiating both sides, we get
	\begin{equation}
	\begin{split}
	\frac{dr^2(t)}{dt} &= \left \langle \frac{d\theta(t)}{dt}, \theta(t) - \theta^* \right \rangle\\
	&=\left \langle -\nabla l(\theta(t)), \theta(t) - \theta^* \right \rangle.
	\end{split}
	\end{equation}
	Using convexity and integrating over all iterates in a trajectory of $T$ time steps,
	\begin{equation}
	\begin{split}
	\frac{1}{T}\int_{0}^{T}\frac{dr^2(t)}{dt} dt & \leq \frac{1}{T}\int_{0}^{T}-\kappa(t)dt\\
	\implies \frac{1}{T}\left ( r^2(T) - r^2(0) \right ) &\leq - \frac{1}{T}\int_{0}^{T}\kappa(t)dt\\
	\implies \frac{1}{T}\int_{0}^{T}\kappa(\theta(t))dt &\leq \frac{r^2(0)}{T}.\\
	\end{split}
	\end{equation}
	By Jensen's inequality,
	\begin{equation}
	\kappa\left ( \frac{1}{T}\int_{0}^{T}\theta(t) dt \right ) \leq \frac{1}{T}\int_{0}^{T}\kappa(\theta(t))dt .
	\end{equation}
	Therefore, $\kappa\left ( \frac{1}{T}\int_{0}^{T}\theta(t) dt \right ) = \mathcal{O}\left (\frac{\left \| \theta(0) - \theta^* \right \|^2}{2T} \right)$  which finishes the proof.\hfill$\square$

	\subsection{Proof of Theorem 5}
	\label{pr_th5}
	In supervised learning with adversarial regularization, the parameters are updated by $\frac{d\theta(t)}{dt} = - \nabla l(\theta(t)) + \nabla g(\theta(t))$. Using arguments of Appendix~\ref{pr_th4}, we obtain
	\begin{equation}
	\begin{split}
	\frac{dr^2(t)}{dt} &=\left \langle -\nabla l(\theta(t)), \theta(t) - \theta^* \right \rangle + \left \langle \nabla g(\theta(t)), \theta(t) - \theta^* \right \rangle.
	\end{split}
	\end{equation}
	Since $l(.)$ is a convex downward and $g(.)$ is a convex upward function, we get
	\begin{equation}
	\begin{split}
	\frac{1}{T}\int_{0}^{T}\frac{dr^2(t)}{dt} dt & \leq -\frac{1}{T}\int_{0}^{T}\kappa(t)dt-\frac{1}{T}\int_{0}^{T}\pi(t)dt \\
	\implies \frac{1}{T}\left ( r^2(T) - r^2(0) \right ) &\leq - \frac{1}{T}\int_{0}^{T}\kappa(t)dt -\frac{1}{T}\int_{0}^{T}\pi(t)dt\\
	\implies \frac{1}{T}\int_{0}^{T}\kappa(\theta(t))dt &\leq \frac{r^2(0)}{T} -\frac{1}{T}\int_{0}^{T}\pi(\theta(t))dt.\\
	\end{split}
	\end{equation}
	Now, using Jensen's inequality on both $\kappa(.)$ and $\pi(.)$
	\begin{equation}
	\kappa\left ( \frac{1}{T}\int_{0}^{T}\theta(t) dt \right ) =\mathcal{O}\left ( \frac{\left \| \theta(0) - \theta^* \right \|^2}{2T} - \pi\left ( \frac{1}{T}\int_{0}^{T}\theta(t) dt \right ) \right)
	\end{equation}
	which finishes the proof.\hfill$\square$
	
	\subsection{Proof of Theorem 6} 
	\label{pr_th6}
	For simplicity, let us denote the bias $b_k = \mathbb{E}\left [ \hat{\mathfrak{g}}_k \right ] - \nabla \mathfrak{l}(\theta_k)$.
	\begin{equation}
	\begin{split}
	\left \| \theta_k  - \theta^* \right \|^2 &= \left \| \theta_{k-1} - \eta_k \hat{\mathfrak{g}}_{k-1} - \theta^* \right \|^2 \\
	& = \left \| \theta_{k-1} - \theta^* \right \|^2 - 2\eta_k \langle \theta_{k-1} - \theta^*, \hat{\mathfrak{g}}_{k-1} \rangle + \eta^2_k\left \| \hat{\mathfrak{g}}_{k-1} \right \|^2\\
	& = \left \| \theta_{k-1} - \theta^* \right \|^2 - 2\eta_k \langle \theta_{k-1} - \theta^*, \nabla \mathfrak{l}(\theta_{k-1}) \rangle - 2\eta_k \langle \theta_{k-1} - \theta^*, b_{k-1} \rangle + \eta^2_k\left \| \hat{\mathfrak{g}}_{k-1} \right \|^2 \\
	& \leq \left \| \theta_{k-1} - \theta^* \right \|^2 - 2\eta_k \langle \theta_{k-1} - \theta^*, \nabla \mathfrak{l}(\theta_{k-1}) \rangle + \underset{\text{By Cauchy-Schwarz inequality}}{\underbrace{2\eta_k \left \| \theta_{k-1} - \theta^* \right \| \left \| b_{k-1} \right \|}} + \eta^2_k\left \| \hat{\mathfrak{g}}_{k-1} \right \|^2 \\
	& \leq \left \| \theta_{k-1} - \theta^* \right \|^2 - 2\eta_k \langle \theta_{k-1} - \theta^*, \nabla \mathfrak{l}(\theta_{k-1}) \rangle + \underset{\text{By AM-GM inequality}}{\underbrace{\eta_k \left ( \left \| \theta_{k-1} - \theta^* \right \|^2 + \left \| b_{k-1} \right \|^2 \right )}} + \eta^2_k\left \| \hat{\mathfrak{g}}_{k-1} \right \|^2
	\end{split}
	\end{equation}
	By $\mu$-strong convexity, it is required that there exist positive constants $\mu$ such that for all $(x,y)$, $\mathfrak{l}(y) \geq \mathfrak{l}(x) + \langle y-x, \nabla \mathfrak{l}(x)\rangle + \frac{\mu}{2} \left \| y-x \right \|^2$. Using strong-convexity at $\theta_{k-1}$ and $\theta^*$, we get
	\begin{equation}
	\begin{split}
	\left \| \theta_k  - \theta^* \right \|^2 & \leq \left \| \theta_{k-1} - \theta^* \right \|^2 - 2\eta_k \left ( \mathfrak{l}(\theta_{k-1}) - \mathfrak{l}(\theta^*) \right ) - \eta_k \mu\left \| \theta_{k-1} - \theta^* \right \|^2 + \eta_k \left ( \left \| \theta_{k-1} - \theta^* \right \|^2 + \left \| b_{k-1} \right \|^2 \right ) + \eta^2_k\left \| \hat{\mathfrak{g}}_{k-1} \right \|^2 \\
	& \leq \left \| \theta_{k-1} - \theta^* \right \|^2 \left ( 1-\eta_k \mu+\eta_k \right ) - 2\eta_k \left ( \mathfrak{l}(\theta_{k-1}) - \mathfrak{l}(\theta^*) \right ) + \eta_k \left \| b_{k-1} \right \|^2 + \eta^2_k\left \| \hat{\mathfrak{g}}_{k-1} \right \|^2.
	\end{split}
	\end{equation}
	
	\textbf{Lemma 3.} \textit{Suppose \textbf{Assumption 7} holds for any $\mathfrak{g}(\theta)$ and $\alpha \in (1,2]$. With global clipping parameter $\tau \geq 0$, the variance and bias of the estimator $\hat{\mathfrak{g}}$ are upper bounded as:
		\begin{equation}
		\mathbb{E}\left [ \left \| \hat{\mathfrak{g}}(\theta) \right \|^2 \right ] \leq G^\alpha \tau^{2-\alpha} \text{and} \left \| \mathbb{E}\left [ \hat{\mathfrak{g}}(\theta) \right ] - \nabla l(\theta) + \nabla g(\theta) \right \|^2 \leq G^{2\alpha} \tau^{2-2\alpha}.
		\end{equation}}
	
	One can easily prove this using \textbf{Lemma 2} of~\cite{zhang2019adam}. Upon rearranging, taking expectation of both sides, and using \textbf{Lemma 3},
	\begin{equation}
	\begin{split}
	\mathbb{E}\left [ \mathfrak{l}(\theta_{k-1}) \right ] - \mathfrak{l}(\theta^*) & \leq \mathbb{E}\left [ \left ( \frac{\eta_k^{-1}-\mu+1}{2} \right ) \left \| \theta_{k-1} - \theta^* \right \|^2  - \frac{\eta_k^{-1}}{2} \left \| \theta_{k} - \theta^* \right \|^2   \right ] + \frac{1}{2} G^{2\alpha} \tau^{2-2\alpha} + \frac{\eta_k}{2}G^\alpha \tau^{2-\alpha}.
	\end{split}
	\end{equation}
	Let us choose $\frac{\eta_k^{-1} - \mu +1}{2} = k-1$ and $\frac{\eta_k^{-1}}{2}=k+1$. After simplification, $\eta_k = \frac{5}{2\mu(k+1)}$. Now, substitute $\tau_k = G k^{\frac{1}{\alpha}} \mu^{\frac{1}{\alpha}}$, $\eta_k = \frac{5}{2\mu(k+1)}$ and multiply $k$ both sides. Thus, 
	\begin{equation}
	\begin{split}
	k\mathbb{E}\left [ \mathfrak{l}(\theta_{k-1}) \right ] - k\mathfrak{l}(\theta^*) &\leq \mathbb{E}\left [ k(k-1) \left \| \theta_{k-1} - \theta^* \right \|^2  - k(k+1) \left \| \theta_{k} - \theta^* \right \|^2   \right ] + \frac{G^2 k^{\frac{2-\alpha}{\alpha}} \mu^{\frac{2-2\alpha}{\alpha}}}{2}\left [ \frac{5}{2}\left ( \frac{k}{k+1} \right )+1 \right ].
	\end{split}
	\end{equation}
	Since $\frac{k}{k+1} < 1$ for $k=1,\dots,T$, we get
	\begin{equation}
	\begin{split}
	k\mathbb{E}\left [ \mathfrak{l}(\theta_{k-1}) \right ] - k\mathfrak{l}(\theta^*) \leq \mathbb{E}\left [ k(k-1) \left \| \theta_{k-1} - \theta^* \right \|^2  - k(k+1) \left \| \theta_{k} - \theta^* \right \|^2   \right ] + \frac{7 G^2 k^{\frac{2-\alpha}{\alpha}} \mu^{\frac{2-2\alpha}{\alpha}}}{4}.
	\end{split}
	\end{equation}
	Taking telescopic sum over $k=1,\dots,T$, we obtain
	\begin{equation}
	\sum_{k=1}^{T}k\mathbb{E}\left [ \mathfrak{l}(\theta_{k-1}) \right ]-\mathfrak{l}(\theta^*)\sum_{k=1}^{T}k \leq \mathbb{E}\left [ -T(T+1)\left \| \theta_T - \theta^* \right \|^2 \right ] + \frac{7G^2 \mu^{\frac{2-2\alpha}{\alpha}}}{4}\sum_{k=1}^{T}k^{\frac{2-\alpha}{\alpha}}.
	\end{equation}
	Using $\sum_{k=1}^{T}k^{\frac{2-\alpha}{\alpha}} \leq \int_{0}^{T+1}k^{\frac{2-\alpha}{\alpha}} dk\leq (T+1)^{\frac{2}{\alpha}}$,
	\begin{equation}
	\sum_{k=1}^{T}k\mathbb{E}\left [ \mathfrak{l}(\theta_{k-1}) \right ]-\mathfrak{l}(\theta^*) \frac{T(T+1)}{2} \leq \frac{7G^2 \mu^{\frac{2-2\alpha}{\alpha}}}{4} (T+1)^{\frac{2}{\alpha}}.
	\end{equation}
	Now, dividing both sides by $\frac{T(T+1)}{2}$ and using $T^{-1} \leq 2(T+1)^{-1}$ for $T\geq 1$,
	\begin{equation}
	\frac{\sum_{k=1}^{T}k\mathbb{E}\left [ \mathfrak{l}(\theta_{k-1}) \right ]}{\sum_{k=1}^{T}k}-\mathfrak{l}(\theta^*) \leq 7G^2 \mu^{\frac{2-2\alpha}{\alpha}} \left ( T+1 \right )^{\frac{2-2\alpha}{\alpha}}.
	\end{equation}
	By Jensen's inequality, 
	\begin{equation}
	\mathbb{E}\left [ \mathfrak{l} \left ( \frac{\sum_{k=1}^{T}k\theta_{k-1}}{\sum_{k=1}^{T}k} \right )\right ]-\mathfrak{l}(\theta^*) \leq \mathcal{O}\left ( G^2 \left ( \mu (T+1) \right )^{\frac{2-2\alpha}{\alpha}} \right )
	\end{equation}
	Substituting $\mathfrak{l} \left( \theta \right ) = l\left ( \theta \right ) - g\left ( \theta \right ) $, we get
	\begin{equation}
	\mathbb{E}\left [ l \left (\bar{\theta} \right )\right ]-l(\theta^*) \leq \mathcal{O}\left ( G^2 \left ( \mu (T+1) \right )^{\frac{2-2\alpha}{\alpha}}  - \left (g \left (\theta^* \right )  - \mathbb{E}\left [ g \left (\bar{\theta} \right )\right ] \right )\right ),
	\end{equation}
	which finishes the proof.\hfill $\square$

	\subsection{Proof of Theorem 7} 
	\label{pr_th7}
	The notations of $\mathfrak{l}$ and $b_k$ follow from Appendix~\ref{pr_th6}. Using $L$-smooth property of $\mathfrak{l}$, we get
	\begin{equation}
	\begin{split}
	\mathfrak{l}\left ( \theta_k \right ) &\leq \mathfrak{l}\left ( \theta_{k-1} \right ) + \langle \nabla \mathfrak{l}\left ( \theta_{k-1} \right ), \theta_k - \theta_{k-1}  \rangle + \frac{L}{2}\left \| \theta_k -\theta_{k-1}\right \|^2\\
	& \leq \mathfrak{l}\left ( \theta_{k-1} \right ) + \langle \nabla \mathfrak{l}\left ( \theta_{k-1} \right ), -\eta_k \hat{\mathfrak{g}}_{k-1} \rangle + \frac{\eta_k^2 L}{2}\left \| \hat{\mathfrak{g}}_{k-1}\right \|^2 \\
	&\leq \mathfrak{l}\left ( \theta_{k-1} \right ) - \eta_k |\left \| \nabla \mathfrak{l}(\theta_{k-1}) \right \|^2 - \eta_k \langle \nabla \mathfrak{l} (\theta_{k-1}), b_{k-1} \rangle + \frac{\eta_k^2 L}{2}\left \| \hat{\mathfrak{g}}_{k-1}\right \|^2 \\
	& \leq \mathfrak{l}\left ( \theta_{k-1} \right ) - \eta_k |\left \| \nabla \mathfrak{l}(\theta_{k-1}) \right \|^2 + \underset{\text{By Cauchy-Schwarz inequality}}{\underbrace{\eta_k  \left \| \nabla \mathfrak{l} (\theta_{k-1}) \right \| \left \| b_{k-1}  \right \|}}  + \frac{\eta_k^2 L}{2}\left \| \hat{\mathfrak{g}}_{k-1}\right \|^2 \\
	&\leq \mathfrak{l}\left ( \theta_{k-1} \right ) - \eta_k |\left \| \nabla \mathfrak{l}(\theta_{k-1}) \right \|^2 + \underset{\text{By AM-GM inequality}}{\underbrace{\frac{\eta_k}{2}  \left ( \left \| \nabla \mathfrak{l} (\theta_{k-1}) \right \|^2 + \left \| \ b_{k-1} \right \|^2 \right )}}  + \frac{\eta_k^2 L}{2}\left \| \hat{\mathfrak{g}}_{k-1}\right \|^2
	\end{split}
	\end{equation}
	Taking expectation of both sides,
	\begin{equation}
	\begin{split}
	\mathbb{E}\left [ \mathfrak{l}(\theta_k) - \mathfrak{l}(\theta_{k-1}) \right ] \leq \mathbb{E}\left [ \frac{-\eta_k}{2} \left \| \nabla \mathfrak{l}(\theta_{k-1}) \right\|^2 \right ] + \frac{\eta_k}{2} G^{2\alpha} \tau^{2-2\alpha} + \frac{\eta_k^2 L}{2} G^\alpha \tau^{2-\alpha}.
	\end{split}
	\end{equation}
	Upon rearranging and taking telescopic sum over $k=1,\dots,T$, we obtain
	\begin{equation}
	\begin{split}
	\frac{1}{T} \sum_{k=1}^{T} \mathbb{E}\left [ \left \| \nabla \mathfrak{l}(\theta_{k-1}) \right\|^2 \right ] \leq \frac{2\eta_k^{-1} }{2}\left (\mathfrak{l}(\theta_0) - \mathfrak{l}(\theta^*)  \right )  +  G^{2\alpha} \tau^{2-2\alpha} + \eta_k L G^\alpha \tau^{2-\alpha}.
	\end{split}
	\end{equation}
	By choosing $\tau = G\left ( \eta_k L \right )^{\frac{-1}{\alpha}}$,
	\begin{equation}
	\begin{split}
	\frac{1}{T} \sum_{k=1}^{T} \mathbb{E}\left [ \left \| \nabla \mathfrak{l}(\theta_{k-1}) \right\|^2 \right ] \leq \frac{2\eta_k^{-1} R_0}{T} + 2 G^2 \left ( \eta_k L \right ) ^{\frac{2\alpha -2}{\alpha}}.
	\end{split}
	\end{equation}
	Let us choose $\eta_k = \left ( \frac{R_0^\alpha L^{2-2\alpha}}{G^2T^\alpha} \right )^{\frac{1}{3\alpha-2}}$. Thus,
	\begin{equation}
	\begin{split}
	\frac{1}{T} \sum_{k=1}^{T} \mathbb{E}\left [ \left \| \nabla \mathfrak{l}(\theta_{k-1}) \right\|^2 \right ] \leq 4 G^{\frac{2\alpha}{3\alpha-2}} \left ( \frac{R_0 L}{T} \right )^{\frac{2\alpha-2}{3\alpha-2}}
	\end{split}
	\end{equation}
	Now, substituting $\mathfrak{l}(\theta) = l(\theta) - g(\theta)$, we get
	\begin{equation}
	\begin{split}
	\frac{1}{T} \sum_{k=1}^{T} \mathbb{E}\left [ \left \| \nabla l(\theta_{k-1}) \right\|^2 + \left \| \nabla g(\theta_{k-1}) \right\|^2 - 2 \langle \nabla l(\theta_{k-1}), \nabla g(\theta_{k-1}) \rangle  \right ] \leq 4 G^{\frac{2\alpha}{3\alpha-2}} \left ( \frac{R_0 L}{T} \right )^{\frac{2\alpha-2}{3\alpha-2}}.
	\end{split}
	\end{equation}
	Since the gradients received from $l(\theta)$ and $g(\theta)$ are negatively correlated at any instant during the optimization process, the above expression simplifies to 
	\begin{equation}
	\begin{split}
	\frac{1}{T} \sum_{k=1}^{T} \mathbb{E}\left [ \left \| \nabla l(\theta_{k-1}) \right\|^2 + \left \| \nabla g(\theta_{k-1}) \right\|^2 + 2 \left \| \nabla l(\theta_{k-1})\right \| \left \|\nabla g(\theta_{k-1}) \right \|  \right ] \leq 4 G^{\frac{2\alpha}{3\alpha-2}} \left ( \frac{R_0 L}{T} \right )^{\frac{2\alpha-2}{3\alpha-2}}.
	\end{split}
	\end{equation}
	Therefore, 
	\begin{equation}
	\begin{split}
	\frac{1}{T} \sum_{k=1}^{T} \mathbb{E}\left [ \left \| \nabla l(\theta_{k-1}) \right\|^2 \right ] + \frac{1}{T} \sum_{k=1}^{T} \mathbb{E}\left [ \left \| \nabla g(\theta_{k-1}) \right\|^2  \right ] \leq 4 G^{\frac{2\alpha}{3\alpha-2}} \left ( \frac{R_0 L}{T} \right )^{\frac{2\alpha-2}{3\alpha-2}}.
	\end{split}
	\end{equation}
	Upon simplification, 
	\begin{equation}
	\begin{split}
	\frac{1}{T} \sum_{k=1}^{T} \mathbb{E}\left [ \left \| \nabla l(\theta_{k-1}) \right\|^2 \right ] &\leq  \mathcal{O}\left (G^{\frac{2\alpha}{3\alpha-2}} \left ( \frac{R_0 L}{T} \right )^{\frac{2\alpha-2}{3\alpha-2}} - \frac{1}{T} \sum_{k=1}^{T} \mathbb{E}\left [ \left \| \nabla g(\theta_{k-1}) \right\|^2  \right ]  \right )
	\end{split}
	\end{equation}
	which finishes the proof. \hfill $\square$

\end{document}